\documentclass{article}
\usepackage{microtype}
\usepackage{graphicx}
\usepackage{subcaption}
\usepackage{booktabs} 
\usepackage{makecell} 
\usepackage[table,xcdraw]{xcolor}
\newcommand{\myrothead}[2][45]{\rotatebox[origin=c]{#1}{\makecell[c]{#2}}}

\usepackage{hyperref}
\usepackage{url}
\usepackage{longtable}
\usepackage{array}
\usepackage{multirow}
\newcolumntype{L}[1]{>{\raggedright\arraybackslash}p{#1}}
\usepackage{amsmath,amssymb,amsfonts}
\usepackage{bm}
\usepackage{amsthm}
\usepackage{mathtools}
\usepackage{algorithm}
\usepackage{algorithmic}
\usepackage{enumitem}
\DeclareUnicodeCharacter{00AD}{}   
\DeclareUnicodeCharacter{FEFF}{}   
\DeclareUnicodeCharacter{FFFE}{}   

\usepackage[preprint]{icml2026}

\usepackage[capitalize,noabbrev]{cleveref}

\theoremstyle{plain}
\newtheorem{theorem}{Theorem}[section]
\newtheorem{proposition}[theorem]{Proposition}
\newtheorem{lemma}[theorem]{Lemma}
\newtheorem{corollary}[theorem]{Corollary}
\theoremstyle{definition}
\newtheorem{definition}[theorem]{Definition}

\theoremstyle{remark}
\newtheorem{remark}[theorem]{Remark}

\usepackage[textsize=tiny]{todonotes}

\icmltitlerunning{Stability of In-Context Learning: A Spectral Coverage Perspective}

\begin{document}

\twocolumn[
  \icmltitle{Stability of In-Context Learning: A Spectral Coverage Perspective}



  \icmlsetsymbol{equal}{*}

  \begin{icmlauthorlist}
    \icmlauthor{Tongxi Wang}{seu}
    \icmlauthor{Zhuoyang Xia}{seu}

  \end{icmlauthorlist}

  \icmlaffiliation{seu}{School of Future Technology, Southeast University, Nanjing, China}

  \icmlcorrespondingauthor{Tongxi Wang}{tongxi\_wang@seu.edu.cn}

  \icmlkeywords{Machine Learning, ICML}

  \vskip 0.3in
]



\printAffiliationsAndNotice{}  

\begin{abstract}
In-context learning (ICL) is a pivotal capability for the practical deployment of large-scale language models, yet its reliability can vary substantially with the number of demonstrations provided in the prompt. A central obstacle is that the target notion, \emph{distributional stability under demonstration resampling}, is expensive to measure directly at scale, making prompt-length selection largely heuristic. We therefore study a \emph{computable sufficient condition} based on a spectral-coverage proxy: the lower tail of the spectrum of a regularized empirical second-moment matrix formed from demonstration representations. Under sub-Gaussian representation assumptions, we derive a non-asymptotic sample-size requirement (a lower bound on $K$) that guarantees this proxy event with prescribed failure probability, yielding a conservative prompt-length recommendation produced by an observable two-stage estimator. In large-scale experiments, the resulting estimates consistently upper-bound empirical accuracy knee-points, which we treat only as a practical surrogate for the prompt-length transition rather than a definition of stability. On a smaller held-out subset, direct resampling-based distributional stability measurements further validate the intended stability interpretation. Finally, a validation-only calibration step tightens the conservatism (typically to about $1.03$--$1.20\times$) while preserving conservative ordering, providing practical and verifiable guidance for ICL prompt design.
\end{abstract}

\begin{figure}[ht]
  \centering
  \includegraphics[width=\linewidth]{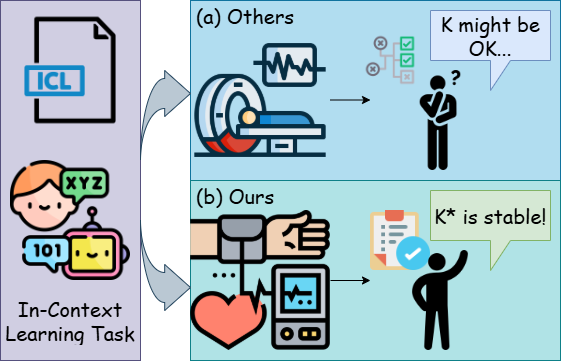}
  \caption{Schematic comparison of \textbf{ICL prompt-length selection}. (a) Conventional trial-and-error or heuristic approaches leave uncertainty about the minimal effective number of demonstrations (K). (b) Our proposed spectral-coverage method provides a theoretically grounded, observable proxy to select a conservative prompt length ($K^\star$) that targets distributional stability under demonstration resampling (Definition~\ref{def:icl-stability}).}
  \label{fig1}
\end{figure}

\section{Introduction}

In-context learning (ICL) has become a key paradigm for large language models to adapt to downstream tasks, as it enables task inference without parameter updates, solely through prompting\cite{wei2022chain,tanwar2023multilingual,dong2024survey}. However, the output of ICL is sensitive to the selection and order of specific examples in the prompt\cite{xie2022explanation,zhao2021calibrate,chen2023relation,li2023finding,fangrethinking}, making its stability difficult to guarantee. An unresolved core question is: \textbf{how can we select a minimal prompt length that yields reliable ICL behavior, when distributional stability under demonstration resampling is costly to measure directly?} The lack of a computable answer to this question leads to reliance on heuristics or excessively long prompts in practice, sacrificing efficiency and failing to provide reliability guarantees.

This paper studies the stability of ICL and the minimal necessary prompt length. We formally define ICL stability in a probabilistic sense: for the same query, two independently sampled sets of examples should induce similar model output distributions. Directly evaluating this stability requires extensive repeated queries to the model, which is costly. Throughout, we treat distributional stability under demonstration resampling as the conceptual target. However, directly estimating it requires repeated model queries for many independently resampled prompts, which is prohibitive for large-scale benchmarking. Our theoretical results therefore concern an \emph{observable spectral-coverage proxy} computed from demonstration representations. Empirically, we evaluate this proxy at scale using accuracy knee-points only as a low-cost surrogate for the prompt-length transition, and we separately validate the proxy--stability connection via direct resampling-based distributional measurements on a smaller subset (Appendix~\ref{app:direct_stability}). To address this, we introduce an observable proxy objective: \textbf{spectral coverage} of example representations, specifically manifested as the minimum eigenvalue of the regularized empirical second-moment (uncentered covariance) matrix. Our goal is to estimate, from data\cite{chang2023data}, the minimum number of examples \(K^*\) required to meet this proxy objective under given stability tolerances \((\tau, \xi)\), and to provide non-asymptotic theoretical guarantees.

In black-box API settings, the generator typically does not expose hidden states.
We therefore instantiate the representation map $\phi$ with a \emph{frozen external encoder} $e(\cdot)$ and compute all spectral quantities in this encoder space.
Our finite-sample guarantees are for this observable proxy (Remark~\ref{rem:proxy-scope}); empirically, we assess proxy--generator alignment via direct resampling-based stability estimates (Appendix~\ref{app:direct_stability}) and a validation-only calibration step (Section~\ref{sec:theory:calibration}).

Previous work has analyzed ICL from various perspectives: theoretical studies have explored learnability under mixture models\cite{chantoward}, sample complexity of Transformers\cite{bai2023transformers}, and asymptotic learning curves\cite{jeon2024information}; calibration methods have improved confidence in few-shot predictions\cite{abbas2024enhancing,cho2025token,tan2025surprise}; mechanistic research has linked internal representations to ICL behavior\cite{dai2023can,hendel2023context}. However, \textbf{none of these works provides a non-asymptotic, computable lower bound for estimating the minimum number of demonstrations required for stable ICL in practical tasks}. Existing rules of thumb are often overly conservative and cannot establish verifiable consistency with the actual behavior of models (e.g., empirical knee-points).

To this end, we propose a framework based on \textbf{spectral coverage theory} to estimate the minimal prompt length required for stable ICL. Our central insight is that \textbf{the spectral lower bound of the empirical second-moment matrix of example representations provides an effective and computable \emph{proxy} for ICL stability}. Specifically, under a linearized ridge-regression view of in-context learning, sufficient spectral coverage controls the conditioning of the induced inverse geometry, thereby limiting the sensitivity of predictions to perturbations in the demonstration set. Building on this insight, we design an \textbf{observable two-stage algorithm}: the first stage uses a small initial sample to estimate spectral statistics and construct a confidence lower bound, while the second stage computes the additional sample size needed to achieve a target level of spectral coverage. Crucially, this procedure preserves non-asymptotic theoretical guarantees without requiring prior knowledge of the population second-moment matrix.

We conduct extensive experiments on multiple tasks covering text classification, multi-step reasoning, and retrieval-augmented generation, involving several commercial API models. 
Our theory yields an explicit \emph{sufficient} sample-size requirement $K \ge K_{\mathrm{th}}$ for the spectral-coverage proxy event (and hence reduced sensitivity to demonstration perturbations in the stylized view). 
Empirically, the resulting estimates $K_{\mathrm{th}}$ also form a conservative upper envelope for observed accuracy knee-points $K_{\mathrm{knee}}$ under fixed decoding protocols. The results show that raw error ratio $\approx 1.3-2.5\times$, defined as the ratio between the predicted prompt length and the empirical knee-point). 
Through a lightweight calibration step using only a validation set (learning a global scaling factor and using quantile eigenvalues), the gap can be significantly tightened (calibrated error ratio $\approx 1.03-1.20\times$). 
We discuss robustness extensions of the assumptions in the appendix for representation drift, heavy-tailed features, and weak dependence.

The contributions of this paper are as follows:

\textbf{Theory (computable proxy with non-asymptotic guarantees):}
    We derive a non-asymptotic \emph{sufficient} condition for achieving a spectral-coverage proxy under sub-Gaussian representations, expressed as an explicit lower bound on the required prompt length $K$. This yields a conservative, fully observable prompt-length recommendation with a prescribed failure probability. We further analyze robustness under representation drift, heavy tails, and weak dependence.
    
\textbf{Observable estimator (no population priors):}
    We design a fully observable two-stage procedure that estimates the required spectral statistics from a small pilot prompt and returns a concrete prompt length with a prescribed failure probability for the proxy event.
    
\textbf{Empirical validation (surrogates at scale, stability checks on a subset):}
    Across datasets, task families, and commercial API models, we show that the proxy-based prompt-length recommendations conservatively upper-bound empirical accuracy knee-points, which we use only as a scalable surrogate for the prompt-length transition. We additionally perform direct resampling-based distributional stability measurements on a smaller held-out subset to validate the intended stability interpretation (Appendix~\ref{app:direct_stability}). A lightweight validation-only calibration (global scale and quantile level) further tightens the conservatism while preserving conservative ordering.

\section{Related Works}
Our work on the minimal prompt length for stable ICL sits at the intersection of three research strands: theoretical analysis of ICL, empirical studies of its reliability, and mechanistic explanations of its emergence. We briefly review these directions and highlight the gap we fill: the lack of a theoretically grounded, practically observable, non-asymptotic, and computable lower bound for ICL stability.

\textbf{Theoretical foundations of ICL.} A growing body of work formalizes ICL within established learning frameworks. PAC-style analyses establish learnability under mixture models, and other studies treat Transformers as algorithms with task-dependent sample complexity\cite{freitrained,magentransformers,li2025transformers}. Linearized asymptotic analyses characterize phase transitions, with task-specific results for regression and structured prediction\cite{chang2025provable,takanami2025learning}. Relatedly, spectrum-oriented post-training interventions aim to improve distributional coverage and in-context steerability\cite{sorensen2025spectrumtuningposttrainingdistributional}. However, these lines generally do not provide non-asymptotic guarantees that yield a concrete demonstration length \(K\) for practical prompt design. We address this by deriving a non-asymptotic lower bound on \(K\) from spectral coverage of demonstration representations, yielding a computable stability proxy.

\textbf{Calibration and reliability in few-shot learning.} Another line of research focuses on calibration methods to improve trustworthiness in ICL predictions. These techniques typically adjust output distributions to enhance predictive confidence and robustness\cite{gundem2025boosting,shen2025exposing}. However, they are orthogonal to our core goal: calibration methods generally assume a fixed prompt and aim to improve the interpretability of its outputs; our goal is to determine the minimal prompt length needed for the output distribution itself to remain stable under demonstration resampling. Thus, our spectral coverage proxy addresses a prerequisite question not tackled by calibration.

\textbf{Mechanistic and empirical perspectives.} Understanding how ICL emerges internally has been explored through mechanistic studies linking Transformer representations to ICL behavior\cite{li2023finding,fangrethinking}. Empirically, benchmarks and toolkits standardize evaluation\cite{chenmaple,rieff2025smmile}, and independent work finds that performance knee-points often require far fewer demonstrations than heuristics suggest\cite{liu2022few,hendel2023context}. Relatedly, in-context clustering work analyzes geometric structure in attention/representations and leverages spectral clustering-style tools in LLM-driven ICL settings\cite{wang2025context}. Complementary to resampling-driven instability, latent shifts have also been studied as a source of ICL unreliability, with self-training proposed as a stabilization mechanism\cite{jukic2024disentangling}. Our theory provides a mechanistic explanation for why accuracy knee-points often appear near the onset of sufficient spectral coverage of demonstration representations, a condition we formalize and make computable.

\section{Theory}\label{sec:theory}

\subsection{Problem setup and proxy objective}\label{sec:theory:setup}

A $K$-shot prompt is formed from $K$ demonstrations $S_K := \{(x_i,y_i)\}_{i=1}^K$ and a query input $x_q$ via a fixed template $\mathrm{Prompt}(\cdot)$.
Given $\mathrm{Prompt}(S_K,x_q)$, a (black-box) language model induces an output distribution
$p_\theta(\cdot \mid \mathrm{Prompt}(S_K,x_q))$ (e.g., over next tokens, or over labels under a verbalizer).
The randomness we study comes from resampling (and, when relevant, re-ordering) the demonstrations in the prompt.

\begin{definition}[ICL stability at $(\tau,\xi)$]\label{def:icl-stability}
Fix a query input $x_q$ and a prompting template.
Let $S_K$ and $S'_K$ be two independent sets of $K$ demonstrations drawn i.i.d.\ from the same task distribution.
Define $P_K := \mathrm{Prompt}(S_K,x_q)$ and $P'_K := \mathrm{Prompt}(S'_K,x_q)$. ICL is stable at length $K$ if
\[
\Pr\!\Big(
d\big(p_\theta(\cdot \mid P_K),\; p_\theta(\cdot \mid P'_K)\big)\le \tau
\Big)\;\ge\; 1-\xi.
\]
where $d(\cdot,\cdot)$ is a divergence between distributions (e.g., total variation, KL, or Jensen--Shannon).
\end{definition}

\paragraph{Intuition and example.}
Definition~\ref{def:icl-stability} formalizes stability in terms of the agreement between output distributions
induced by two independently sampled prompts of the same length.
Concretely, consider a classification task where each prompt consists of $K$ labeled demonstrations.
Two prompts $P_K$ and $P'_K$ are formed by independently sampling and ordering demonstrations from the same task pool.
If the model is \emph{unstable} at length $K$, small changes in the selected demonstrations may lead to
noticeably different output distributions---for example, flipping the predicted label or substantially
redistributing probability mass among competing classes.
In contrast, stability implies that, despite variation in the demonstrations, the model produces
highly similar predictive distributions, reflecting insensitivity to the specific realization of the prompt.
This notion captures robustness to prompt composition rather than average accuracy, and is task-agnostic.

\begin{remark}[Scope of the proxy and instantiation of $\phi$]\label{rem:proxy-scope}
Definition~\ref{def:icl-stability} concerns the LLM output distribution, but our analysis uses an \emph{observable} proxy
computed from a representation map $\phi:\mathcal{X}\to\mathbb{R}^d$.
In black-box API settings we instantiate $\phi$ with a frozen external encoder and compute all spectral quantities in this
encoder space; implementation details are in Appendix~\ref{app:repro}.
\end{remark}

Directly verifying Definition~\ref{def:icl-stability} is expensive because it requires repeated model queries under many resampled prompts.
We therefore study an \emph{observable} sufficient condition based on the spectral coverage of demonstration representations.
Let $\phi:\mathcal{X}\to\mathbb{R}^d$ be a fixed feature map and define the (uncentered) population second-moment matrix
\[
\Sigma := \mathbb{E}\!\left[\phi(x)\phi(x)^\top\right]\succeq 0,
\]
which does not require $\mathbb{E}[\phi(x)]=0$.
Given i.i.d.\ demonstrations $x_1,\ldots,x_K$, define
\begin{align}    
H_K &:=
\begin{bmatrix}
\phi(x_1)^\top\\
\vdots\\
\phi(x_K)^\top
\end{bmatrix}\in\mathbb{R}^{K\times d}, \\
\widehat{\Sigma}_K &:= \frac{1}{K}H_K^\top H_K
= \frac{1}{K}\sum_{k=1}^K X_k,\;\;\; X_k:=\phi(x_k)\phi(x_k)^\top.\notag 
\end{align}
For numerical robustness we use ridge regularization: for a ridge level $\rho\ge 0$, let
\[
\widehat{\Sigma}_{K,\rho} := \widehat{\Sigma}_K + \rho I_d.
\]
Our primary proxy objective is to find the minimal $K^\star$ such that
\begin{equation}\label{eq:spectral-floor-prob}
\Pr\!\big(\lambda_{\min}(\widehat{\Sigma}_{K,\rho}) \ge \delta\big)\;\ge\; 1-\xi,
\end{equation}
for a target floor $\delta>\rho$ and failure tolerance $\xi\in(0,1)$.
This proxy is equivalent to requiring $\lambda_{\min}(\widehat{\Sigma}_K)\ge \delta-\rho$ but is more stable numerically and aligns with ridge/regularized linearized views of ICL.

To reduce sensitivity to the extreme tail of the spectrum, we will also consider a quantile surrogate.
Let $\lambda^{(1)}(A)\le\cdots\le \lambda^{(d)}(A)$ denote the eigenvalues of a PSD matrix $A$ in nondecreasing order, and define
\[
\lambda_q(A) := \lambda^{(\lceil qd\rceil)}(A),\qquad q\in(0,1).
\]
In practice we often use $\lambda_q(\widehat{\Sigma}_{K,\rho})$ (e.g., $q\in[0.05,0.2]$) to obtain less conservative yet still monotone proxies.
Throughout the main analysis we adopt the following standard nominal condition on $\phi(x)$.

(A1)\textbf{Sub-Gaussian features.}The feature vector $\phi(x)\in\mathbb{R}^d$ is sub-Gaussian (e.g., $\|\phi(x)\|_{\psi_2}\le \sigma$) and $\Sigma=\mathbb{E}[\phi(x)\phi(x)^\top]$ exists.

This assumption should be viewed as a working condition enabling clean non-asymptotic analysis; in the appendix, we discuss extensions to heavy-tailed features, representation drift, weak dependence, and empirically evaluate robustness beyond this idealized setting.

We discuss robustness extensions under representation drift, heavier tails, and weak dependence in the appendix (moved from the original main-text derivations).

\textbf{Notation.}
Let $r=\mathrm{rank}(\Sigma)$ and denote the smallest nonzero eigenvalue by $\lambda_r(\Sigma)>0$; for brevity we write $\lambda_{\min}(\Sigma)=\lambda_r(\Sigma)$.
We use $\|\cdot\|$ for the operator norm and $\mathrm{tr}(\cdot)$ for trace.
``Effective rank'' will be measured by
\[
r_{\mathrm{eff}} := \frac{\mathrm{tr}(\Sigma)}{\|\Sigma\|} \le r,
\]
which naturally appears in matrix concentration bounds.

\subsection{Sample complexity bound}
\label{sec:theory:main}

We study the proxy objective in~\eqref{eq:spectral-floor-prob} and derive a non-asymptotic sample complexity guaranteeing a spectral ``floor'' for the ridge-regularized empirical second moment.
The full derivation (matrix Bernstein lower tail, truncation, and variance proxy control) is moved to the appendix; here we state only the final, implementation-facing bound.

Define the spectral margin
\begin{equation}\label{eq:Delta_rho_def}
\Delta_\rho \;:=\; \lambda_{\min}(\Sigma) + \rho - \delta.
\end{equation}
The regime of interest is $\Delta_\rho>0$, i.e., the target floor $\delta$ is below the population ridge floor $\lambda_{\min}(\Sigma)+\rho$.

\begin{proposition}[Spectral coverage from $K$ demonstrations]\label{prop:spectral_coverage_bound}
Assume (A1) (sub-Gaussian features) and let $\xi\in(0,1)$.
There exist universal constants $c,C>0$ such that if $\Delta_\rho>0$ and
\begin{equation}\label{eq:K_main_simpler}
K \;\ge\; C \cdot \frac{\|\Sigma\|^2}{\Delta_\rho^2}\, r_{\mathrm{eff}}\,
\log\!\Big(\frac{c\, r_{\mathrm{eff}}}{\xi}\Big),
\end{equation}
then the proxy objective holds:
\begin{equation}\label{eq:spectral_floor_success}
\Pr\!\big(\lambda_{\min}(\widehat{\Sigma}_{K,\rho}) \ge \delta\big)\;\ge\; 1-\xi.
\end{equation}
\end{proposition}

The leading dependence in~\eqref{eq:K_main_simpler} is the familiar
\[
K \;\asymp\; \frac{\text{(scale)}^2 \times \text{(effective dimension)} \times \log(1/\xi)}{\text{(spectral margin)}^2},
\]
where the scale is $\|\Sigma\|$, the effective dimension is $r_{\mathrm{eff}}=\mathrm{tr}(\Sigma)/\|\Sigma\|$, and the margin is $\Delta_\rho$.
Intuitively, larger feature variance (larger $\|\Sigma\|$) and higher intrinsic dimensionality (larger $r_{\mathrm{eff}}$) require more demonstrations to ensure that the empirical covariance is well-conditioned along \emph{all} directions, while a larger margin $\Delta_\rho$ makes the guarantee easier to satisfy.

The appendix also provides an explicit non-asymptotic sufficient condition with additional higher-order and logarithmic terms (arising from controlling per-sample operator norms and a self-consistency step).
We omit that longer expression in the main text to keep the theory readable; see the appendix for the complete chain and constants.

Although Proposition~\ref{prop:spectral_coverage_bound} is stated in terms of population quantities, it is designed to be used via plug-in estimates of
$\|\Sigma\|$, $r_{\mathrm{eff}}$, and $\lambda_{\min}(\Sigma)$.
Section~\ref{sec:theory:twostage} introduces an observable two-stage procedure that estimates these terms from data and returns a concrete prompt length $K$ with the desired failure probability.

\subsection{An observable two-stage estimator}\label{sec:theory:twostage}

Proposition~\ref{prop:spectral_coverage_bound} suggests choosing $K$ using population quantities
$\lambda_{\min}(\Sigma)$, $\|\Sigma\|$, and $r_{\mathrm{eff}}$.
In practice these are unknown, and we seek an \emph{observable} procedure that (i) estimates the needed scale and
intrinsic dimension from data, and (ii) returns a concrete prompt length with a prescribed failure probability.

We use a small pilot prompt to obtain a coarse but reliable estimate of the spectrum, and then plug these estimates
into the readable bound~\eqref{eq:K_main_simpler}.
To guard against optimism from finite-sample noise, we include a safety margin on the spectral gap~\eqref{eq:Delta_rho_def}
and allocate the total failure probability across the two stages.

Given $K_0$ demonstrations, form $\widehat{\Sigma}_{K_0}$ and $\widehat{\Sigma}_{K_0,\rho}=\widehat{\Sigma}_{K_0}+\rho I$ as in
Section~\ref{sec:theory:setup}, and compute the observable quantities
\begin{equation}\label{eq:pilot_stats}
\widehat{\lambda}_0:=\lambda_{\min}(\widehat{\Sigma}_{K_0}),
\widehat{s}_0:=\|\widehat{\Sigma}_{K_0}\|,
\widehat{r}_{\mathrm{eff},0}:=\frac{\mathrm{tr}(\widehat{\Sigma}_{K_0})}{\|\widehat{\Sigma}_{K_0}\|}.
\end{equation}
We then define a conservative estimated margin
\begin{equation}\label{eq:Delta_hat_def}
\widehat{\Delta}_\rho \;:=\; \big(\widehat{\lambda}_0 - \eta\big) + \rho - \delta,
\end{equation}
where $\eta\ge 0$ is a user-chosen slack (or a bound-driven correction; see appendix for one principled choice).
If $\widehat{\Delta}_\rho\le 0$, we increase $K_0$ and repeat.

Once $\widehat{\Delta}_\rho>0$, we output a plug-in length
\begin{equation}\label{eq:K_hat_plugin}
\widehat{K}\;:=\;
\left\lceil
C \cdot \frac{\widehat{s}_0^{\,2}}{\widehat{\Delta}_\rho^{\,2}}\,
\widehat{r}_{\mathrm{eff},0}\,
\log\!\Big(\frac{c\,\widehat{r}_{\mathrm{eff},0}}{\xi_2}\Big)
\right\rceil,
\end{equation}
where $\xi_2$ is the failure budget assigned to Stage~II and $c,C$ are the same universal constants as in
Proposition~\ref{prop:spectral_coverage_bound}.
In experiments we optionally apply a single global calibration factor to reduce conservatism (Section on calibration).

\begin{algorithm}[t]
\caption{Two-stage observable estimator for the minimal prompt length}
\label{alg:two-stage}
\begin{algorithmic}[1]
\REQUIRE tolerance $\xi\in(0,1)$; target $\delta$; ridge coefficient $\epsilon_\rho>0$; initial $K_0$ (e.g., $K_0=100$)
\STATE $j \gets 0$
\REPEAT
  \STATE $\xi_j \gets \xi/2^{j+2}$ 
    \COMMENT{geometric allocation; $\sum_{j\ge 0}\xi_j \le \xi/2$}
  \STATE Draw $K_0$ demonstrations and compute $\widehat{\Sigma}_{K_0}$
  \STATE $\rho \gets \min\{\epsilon_\rho\cdot \mathrm{tr}(\widehat{\Sigma}_{K_0})/d,\ \delta/2\}$
  \STATE $\widehat{\Sigma}_{K_0,\rho} \gets \widehat{\Sigma}_{K_0}+\rho I_d$
  \STATE $\widehat{\lambda}_0 \gets \lambda_{\min}(\widehat{\Sigma}_{K_0,\rho})$
  \STATE $\widehat{\|\Sigma\|} \gets \|\widehat{\Sigma}_{K_0}\|$
  \STATE $\widehat{r}_{\mathrm{eff}} \gets \mathrm{tr}(\widehat{\Sigma}_{K_0})/\|\widehat{\Sigma}_{K_0}\|$
  \STATE \textbf{Lower confidence bound}:
  \STATE $\underline{\lambda} \gets \widehat{\lambda}_0 -
    C\,\widehat{\|\Sigma\|}\!\left(
      \sqrt{\tfrac{\widehat{r}_{\mathrm{eff}}\log(2/\xi_j)}{K_0}} +
      \tfrac{\widehat{r}_{\mathrm{eff}}\log(2/\xi_j)}{K_0}
    \right)$
  \IF{$\underline{\lambda} \le 0$}
    \STATE $K_0 \gets 2K_0$; $j \gets j+1$
  \ENDIF
\UNTIL{$\underline{\lambda} > 0$}
\STATE $\widehat{\Delta} \gets \underline{\lambda} - \delta$
\IF{$\widehat{\Delta} \le 0$}
  \STATE \textbf{return} \textbf{infeasible target}
\ENDIF
\STATE $K \gets C'\,\widehat{\|\Sigma\|}^2\,\widehat{r}_{\mathrm{eff}}
  \log(4/\xi)\,/\,\widehat{\Delta}^2$
\STATE \textbf{return} $K_{\mathrm{final}} \gets K_0 + K$
\end{algorithmic}
\end{algorithm}

Algorithm~\ref{alg:two-stage} is fully observable given a feature map $\phi$:
it requires only computing Gram/covariance statistics of demonstration representations.
If it returns \textbf{infeasible target}, then the requested floor $\delta$ exceeds the current lower confidence bound $\underline{\lambda}$ (i.e., $\widehat{\Delta}\le 0$), indicating that the chosen target is not supported at confidence $1-\xi$ under the current feature map and ridge level; practical remedies and calibration guidance are given in Appendix~\ref{app:const}.
The appendix contains a complete non-asymptotic analysis showing how to choose $K_0$ and $\eta$
(and how to split $\xi$) so that the returned $\widehat{K}$ attains the target guarantee in~\eqref{eq:spectral_floor_success}
under the same assumptions as Proposition~\ref{prop:spectral_coverage_bound}.

Both stages operate solely on the demonstration pool (or training split) and do not access
test-query outputs or labels; the resulting prompt length is fixed prior to evaluation.

\subsection{Spectral floor as stability proxy}
\label{sec:theory:icl-connection}

The proxy in~\eqref{eq:spectral-floor-prob} is purely geometric: it asks whether the demonstration features span the space
well enough so that the empirical second moment is well-conditioned.
This section provides a modeling rationale---via ridge-regularized linearization---for using such a spectral ``floor'' as a surrogate for ICL stability, and clarifies what the proxy does \emph{and does not} capture.
We complement the theory with empirical diagnostics in Appendix~\ref{app:interpret} and direct stability measurements in Appendix~\ref{app:direct_stability}.

Many analyses of ICL posit that, for a fixed query $x_q$, the model's output depends on demonstrations primarily through a
low-dimensional statistic built from representations (e.g., attention-weighted averages, or a fitted linear probe).
A generic abstraction is:
(i) the prompt induces a parameter estimate $\widehat{w}$ by solving a regularized least-squares problem in feature space,
and (ii) the model output at the query is a Lipschitz function of $\widehat{w}$.
Under this view, stability reduces to controlling how much $\widehat{w}$ varies when we resample the demonstrations.

Let $H_K\in\mathbb{R}^{K\times d}$ be the feature matrix defined in Section~\ref{sec:theory:setup}.
Consider the ridge estimator
\begin{align}\label{eq:ridge_estimator}
\widehat{w}(S_K) \;:&=\; \arg\min_{w\in\mathbb{R}^d}\;
\frac{1}{K}\|H_K w - y\|_2^2 + \rho \|w\|_2^2 \notag \\
\;&=\;
\big(\widehat{\Sigma}_{K,\rho}\big)^{-1}\,\Big(\frac{1}{K}H_K^\top y\Big),
\end{align}
where $y\in\mathbb{R}^K$ denotes supervision associated with the demonstrations (e.g., class labels under a fixed
encoding, or pseudo-labels produced by the model itself), and $\rho\ge 0$ is the same ridge level as in~\eqref{eq:spectral-floor-prob}.

\paragraph{Why spectral coverage implies stability.}
At a high level, the smallest eigenvalue of the empirical second-moment matrix
$\widehat{\Sigma}_K = \frac{1}{K}\sum_{i=1}^K \phi(x_i)\phi(x_i)^\top$
quantifies how well the representation vectors of the demonstrations span the feature space.
When $\lambda_{\min}(\widehat{\Sigma}_K)$ is small, the induced inverse geometry is poorly conditioned:
small perturbations in the empirical moments---arising from changing or resampling a few demonstrations---can be greatly amplified,
leading to large variations in the fitted coefficients and, consequently, in the model's predictions.
In contrast, a larger spectral lower bound ensures that all directions in representation space are sufficiently covered,
which stabilizes the inverse problem and suppresses amplification effects.
This intuition extends to quantile-based spectral surrogates, which trade off strict worst-case control
for improved numerical robustness in finite samples.

\begin{lemma}[Spectral floor $\Rightarrow$ bounded sensitivity of ridge parameters]\label{lem:spectral-to-stability}
Assume $\|y\|_2\le B\sqrt{K}$ almost surely for some $B>0$.
Let $S_K$ and $S'_K$ be two independent $K$-demo prompts with feature matrices $H_K,H'_K$
and ridge covariances $\widehat{\Sigma}_{K,\rho},\widehat{\Sigma}'_{K,\rho}$.
On the event
\[
\mathcal{E}_\delta \;:=\;
\Big\{\lambda_{\min}(\widehat{\Sigma}_{K,\rho})\ge \delta\Big\}
\;\cap\;
\Big\{\lambda_{\min}(\widehat{\Sigma}'_{K,\rho})\ge \delta\Big\},
\]
the ridge solutions satisfy the deterministic bound
\begin{equation}\label{eq:ridge_param_sensitivity}
\|\widehat{w}(S_K)-\widehat{w}(S'_K)\|_2
\;\le\;
\frac{2B}{\delta}\;+\;\frac{B}{\delta^2}\,\|\widehat{\Sigma}_{K}-\widehat{\Sigma}'_{K}\| \, .
\end{equation}
In particular, conditioning improves as the spectral floor $\delta$ increases.
\end{lemma}

\textbf{From ridge sensitivity to distributional stability (sketch).}
Lemma~\ref{lem:spectral-to-stability} shows that a spectral floor
$\lambda_{\min}(\widehat{\Sigma}_{K,\rho})\ge \delta$ prevents the ridge inverse from amplifying
prompt-to-prompt variability in the induced ridge parameter $\widehat w$.
Under a mild Lipschitz dependence of the model output distribution on $\widehat w$,
this translates into small divergence between $p_\theta(\cdot\mid P_K)$ across independently resampled prompts,
with probability controlled by the proxy failure level $\xi$.
We provide the full derivation (including the Lipschitz condition, intermediate inequalities, and sharper variants
that avoid explicit $\|\widehat{\Sigma}_K-\widehat{\Sigma}'_K\|$ terms) in Appendix~\ref{app:interpret}.

\subsection{Calibration and practical recommendations}\label{sec:theory:calibration}

The bound in Proposition~\ref{prop:spectral_coverage_bound} and the plug-in rule~\eqref{eq:K_hat_plugin}
are intentionally conservative: they protect against worst-case directions in the feature space and use
high-probability concentration. Importantly, calibration is introduced as a practical correction for proxy--generator mismatch and finite-sample conservatism; it does not change the non-asymptotic guarantee for the uncalibrated proxy bound, and should be viewed as an empirical tightening step rather than a first-principles refinement of the theory.
In practice, two issues commonly arise:
(i) the smallest eigenvalue is overly sensitive to noise in near-null directions, and
(ii) constants hidden in non-asymptotic inequalities can inflate the suggested prompt length.
This section summarizes lightweight calibration strategies that preserve the monotonic behavior of the proxy
while producing practically useful $K$.

Instead of targeting the extreme floor $\lambda_{\min}$, it is often more stable to track the quantile eigenvalue
$\lambda_q(\widehat{\Sigma}_{K,\rho})$ defined in Section~\ref{sec:theory:setup}.
Operationally, one may replace the event in~\eqref{eq:spectral-floor-prob} by
\begin{equation}\label{eq:quantile_proxy_prob}
\Pr\!\big(\lambda_q(\widehat{\Sigma}_{K,\rho}) \ge \delta_q\big)\;\ge\; 1-\xi,
\end{equation}
for a chosen $q\in(0,1)$ (e.g., $q\in[0.05,0.2]$) and threshold $\delta_q$.
This reduces sensitivity to a small number of poorly-covered directions while still enforcing broad spectral coverage.
The same two-stage estimator (Algorithm~\ref{alg:two-stage}) applies by replacing $\widehat{\lambda}_0$ in~\eqref{eq:pilot_stats}
with $\lambda_q(\widehat{\Sigma}_{K_0})$ and $\Delta_\rho$ in~\eqref{eq:Delta_rho_def} with the corresponding quantile margin.

Let $\widehat{K}$ denote the plug-in output from~\eqref{eq:K_hat_plugin}.
We optionally apply a global calibration factor $\alpha\in(0,1]$ and use
\begin{equation}\label{eq:K_calibrated}
\widehat{K}_{\mathrm{cal}} \;:=\; \left\lceil \alpha\,\widehat{K} \right\rceil.
\end{equation}
A practical alternative (used in our experiments) is to calibrate $(q,\alpha)$ against empirical knee-points
estimated on training data only (no test leakage; Appendix~\ref{app:repro}).
For each encoder--generator pair, we select a small set of calibration tasks and estimate
$\widehat K_{\mathrm{knee}}^{(t)}$ on a held-out validation subset drawn from the training split.
We then choose $q^\star \in [0.05,0.2]$ by minimizing the median absolute log-error
between $\widehat K_{\mathrm{raw}}^{(t)}(q)$ and $\widehat K_{\mathrm{knee}}^{(t)}$ across calibration tasks, and set
$\alpha$ by a robust median ratio fit:
\[
\alpha \;:=\; \mathrm{median}_{t}\ \widehat K_{\mathrm{knee}}^{(t)}\big/\widehat K_{\mathrm{raw}}^{(t)}(q^\star).
\]
Finally, we freeze $(q^\star,\alpha)$ and report both raw and calibrated predictions on disjoint evaluation tasks.

\paragraph{Default practical choices.}
Robust default settings (ridge level, spectral threshold choice, and failure-budget split) are summarized in Appendix~\ref{app:const:defaults}.

\begin{table*}[t]
\centering
\small
\setlength{\tabcolsep}{2pt}
\renewcommand{\arraystretch}{1.05}
\caption{
Error ratio ($K^*/\text{knee}$) across datasets and task families for commercial API generators.
Encoder groups a/b/c correspond to \textbf{all-mpnet-base-v2}, \textbf{all-MiniLM-L6-v2}, and
\textbf{paraphrase-MiniLM-L12-v2}. Entries report the mean ratio over $5$ random seeds; per-cell standard deviations
(and per-seed raw outputs) are provided in the released artifact and summarized in Appendix~\ref{app:repro}.
Knee-points are estimated on the test split via a two-segment piecewise-linear fit on the grid $\mathcal{K}$
with bootstrap confidence intervals (Appendix~\ref{app:repro}).
The Raw column uses the default theoretical constants, whereas Cal.\ uses validation-only calibration
(Appendix~\ref{app:para}). Lower is better (closer to~1).
}
\resizebox{\textwidth}{!}{%
\begin{tabular}{l l c c c c c c c c c c c c}
\toprule
\multirow{2}{*}{Dataset} & \multirow{2}{*}{Enc.} &
\multicolumn{2}{c}{\myrothead[45]{OpenAI\\gpt-5.2-chat-latest}} &
\multicolumn{2}{c}{\myrothead[45]{Anthropic\\claude-sonnet-4-5}} &
\multicolumn{2}{c}{\myrothead[45]{Gemini\\gemini-3-flash-preview}} &
\multicolumn{2}{c}{\myrothead[45]{DeepSeek\\deepseek-reasoner}} &
\multicolumn{2}{c}{\myrothead[45]{Kimi\\kimi-k2-thinking}} &
\multicolumn{2}{c}{\myrothead[45]{Qwen\\qwen-max-latest}} \\
\cmidrule(lr){3-4}\cmidrule(lr){5-6}\cmidrule(lr){7-8}\cmidrule(lr){9-10}\cmidrule(lr){11-12}\cmidrule(lr){13-14}
& & Raw & Cal. & Raw & Cal. & Raw & Cal. & Raw & Cal. & Raw & Cal. & Raw & Cal. \\
\midrule

\rowcolor{green!8} SST-2 & a & 1.46$\times$ & 1.06$\times$ & 1.43$\times$ & 1.03$\times$ & 1.56$\times$ & 1.07$\times$ & 1.39$\times$ & 1.05$\times$ & 1.47$\times$ & 1.04$\times$ & 1.51$\times$ & 1.07$\times$ \\
\rowcolor{green!8}  & b & 1.35$\times$ & 1.02$\times$ & 1.31$\times$ & 1.04$\times$ & 1.45$\times$ & 1.06$\times$ & 1.29$\times$ & 1.04$\times$ & 1.33$\times$ & 1.03$\times$ & 1.39$\times$ & 1.06$\times$ \\
\rowcolor{green!8}  & c & 1.41$\times$ & 1.05$\times$ & 1.37$\times$ & 1.03$\times$ & 1.51$\times$ & 1.05$\times$ & 1.35$\times$ & 1.05$\times$ & 1.39$\times$ & 1.06$\times$ & 1.45$\times$ & 1.05$\times$ \\
\midrule

\rowcolor{orange!10} AGNews & a & 1.77$\times$ & 1.09$\times$ & 1.69$\times$ & 1.06$\times$ & 1.85$\times$ & 1.10$\times$ & 1.63$\times$ & 1.08$\times$ & 1.75$\times$ & 1.07$\times$ & 1.81$\times$ & 1.11$\times$ \\
\rowcolor{orange!10}  & b & 1.63$\times$ & 1.08$\times$ & 1.57$\times$ & 1.05$\times$ & 1.71$\times$ & 1.07$\times$ & 1.49$\times$ & 1.07$\times$ & 1.61$\times$ & 1.08$\times$ & 1.69$\times$ & 1.10$\times$ \\
\rowcolor{orange!10}  & c & 1.69$\times$ & 1.08$\times$ & 1.61$\times$ & 1.06$\times$ & 1.77$\times$ & 1.08$\times$ & 1.55$\times$ & 1.08$\times$ & 1.67$\times$ & 1.09$\times$ & 1.75$\times$ & 1.09$\times$ \\
\midrule

\rowcolor{red!8} {Yahoo! Answers} & a & 1.69$\times$ & 1.09$\times$ & 1.63$\times$ & 1.06$\times$ & 1.77$\times$ & 1.08$\times$ & 1.53$\times$ & 1.08$\times$ & 1.65$\times$ & 1.09$\times$ & 1.73$\times$ & 1.11$\times$ \\
\rowcolor{red!8}  & b & 1.55$\times$ & 1.07$\times$ & 1.47$\times$ & 1.06$\times$ & 1.63$\times$ & 1.07$\times$ & 1.39$\times$ & 1.06$\times$ & 1.51$\times$ & 1.07$\times$ & 1.59$\times$ & 1.09$\times$ \\
\rowcolor{red!8}  & c & 1.61$\times$ & 1.08$\times$ & 1.53$\times$ & 1.07$\times$ & 1.69$\times$ & 1.07$\times$ & 1.45$\times$ & 1.07$\times$ & 1.57$\times$ & 1.08$\times$ & 1.65$\times$ & 1.08$\times$ \\
\midrule

\rowcolor{blue!8} TREC-6 & a & 1.43$\times$ & 1.05$\times$ & 1.39$\times$ & 1.04$\times$ & 1.49$\times$ & 1.06$\times$ & 1.33$\times$ & 1.04$\times$ & 1.41$\times$ & 1.05$\times$ & 1.47$\times$ & 1.06$\times$ \\
\rowcolor{blue!8}  & b & 1.35$\times$ & 1.04$\times$ & 1.31$\times$ & 1.02$\times$ & 1.41$\times$ & 1.05$\times$ & 1.27$\times$ & 1.03$\times$ & 1.33$\times$ & 1.04$\times$ & 1.39$\times$ & 1.05$\times$ \\
\rowcolor{blue!8}  & c & 1.39$\times$ & 1.04$\times$ & 1.35$\times$ & 1.04$\times$ & 1.45$\times$ & 1.03$\times$ & 1.31$\times$ & 1.04$\times$ & 1.37$\times$ & 1.04$\times$ & 1.43$\times$ & 1.06$\times$ \\

\midrule
\rowcolor{gray!12}
\textbf{Task family} & \textbf{Dataset} & Raw & Cal. & Raw & Cal. & Raw & Cal. & Raw & Cal. & Raw & Cal. & Raw & Cal. \\
\midrule

\rowcolor{purple!8} Reasoning & GSM8K
& 2.03$\times$ & 1.14$\times$
& 1.91$\times$ & 1.12$\times$
& 2.22$\times$ & 1.16$\times$
& 1.87$\times$ & 1.10$\times$
& 1.90$\times$ & 1.11$\times$
& 2.14$\times$ & 1.13$\times$ \\
\rowcolor{teal!8} RAG & HotpotQA
& 2.33$\times$ & 1.18$\times$
& 2.24$\times$ & 1.16$\times$
& 2.50$\times$ & 1.20$\times$
& 2.09$\times$ & 1.14$\times$
& 2.16$\times$ & 1.15$\times$
& 2.31$\times$ & 1.18$\times$ \\

\bottomrule
\end{tabular}%
}
\label{tab:main_ratio}
\end{table*}

\section{Experiments}

We conduct experiments to validate the theoretical lower bound on the minimal prompt length required for stable In-Context Learning (ICL). The experiments test the predictions of our theory across various datasets, encoders, and generators. Unless otherwise stated, we use deterministic decoding (e.g., temperature $=0$ when supported) so that variability is dominated by demonstration resampling rather than token-level sampling noise. Full reproducibility specifications (prompt templates, demo selection/retrieval, representation extraction, decoding hyperparameters, random seeds, knee-point estimator, and confidence intervals) are provided in Appendix~\ref{app:repro}.

\subsection{Experimental setup}
We provide a comprehensive experimental inventory to facilitate reproduction, ablation and future benchmarking:

\textbf{Datasets:} SST-2\cite{socher2013recursive}, AGNews\cite{Zhang2015CharacterlevelCN}, Yahoo!Answers\cite{Zhang2015CharacterlevelCN}, TREC-6\cite{trec6}, GSM8K\cite{cobbe2021training}, SVAMP\cite{patel2021nlp}, HotpotQA\cite{yang2018hotpotqa} and Natural Questions\cite{kwiatkowski2019natural}, covering sentiment, news topic, QA and multi-step reasoning.

\textbf{Encoders:} all-mpnet-base-v2, all-MiniLM-L6-v2 and paraphrase-MiniLM-L12-v2\cite{reimers2019sentence,song2020mpnet,wang2021minilmv2}.

\textbf{Models:} Anthropic Claude-Sonnet-4-5, DeepSeek-Reasoner\cite{guo2025deepseek}, Gemini-3-Flash-Preview, Kimi-k2-thinking\cite{team2025kimi}, OpenAI gpt-5.2-chat-latest and Qwen-Max-Latest\cite{yang2025qwen3}.

\begin{table}[ht]
\centering
\caption{Numerical noise sensitivity of predicted prompt length $K^*$.
Lower CV indicates higher numerical robustness.}
\label{tab:numerical-sensitivity}
\begin{tabular}{lccc}
\toprule
Noise level $\varepsilon$ &
\myrothead[0]{CV($K^*$ \\ via $\lambda_{\min}$)} &
\myrothead[0]{CV($K^*$ \\ via $\lambda_{\min}+\rho I$)} &
\myrothead[0]{CV($K^*$ \\ via $\lambda_q$)} \\
\midrule
$10^{-6}$ & 0.09 & 0.012 & 0.009 \\
$10^{-5}$ & 0.21 & 0.024 & 0.017 \\
$10^{-4}$ & 0.47 & 0.071 & 0.052 \\
$10^{-3}$ & 0.98 & 0.19  & 0.13  \\
\bottomrule
\end{tabular}
\end{table}

\subsection{Measuring the Knee-Point}
\label{sec:knee}
For each (task, encoder, generator) setting, we estimate an empirical saturation point
$\widehat K_{\mathrm{knee}}$ from an accuracy--vs.--prompt-length curve evaluated on a discrete grid
$\mathcal{K}$.
Concretely, for each $K\in\mathcal{K}$ we measure $\mathrm{ACC}(K)$ on the test split under a fixed
demonstration construction protocol (including retrieval, formatting, and decoding).
We then fit a two-segment piecewise-linear model to $\{(K,\mathrm{ACC}(K))\}_{K\in\mathcal{K}}$
and define $\widehat K_{\mathrm{knee}}$ as the breakpoint minimizing the sum of squared errors (SSE),
with ties broken toward smaller $K$ for conservatism.
To quantify uncertainty, we compute a percentile bootstrap confidence interval for $\widehat K_{\mathrm{knee}}$
by resampling evaluation queries with replacement (Appendix~\ref{app:repro}).

Although $\widehat K_{\mathrm{knee}}$ is defined via accuracy saturation rather than Definition~\ref{def:icl-stability}, we use it only as an empirical baseline for comparison. A plateau in accuracy alone does not certify distributional stability, particularly under stringent stability tolerances or open-generation settings. To support the intended stability interpretation, we additionally measure distributional stability under repeated demonstration resampling on a held-out subset; see Appendix~\ref{app:direct_stability}.

We summarize prediction quality using the \emph{error ratios}
\[
R_{\mathrm{raw}} := \frac{K^*}{\widehat K_{\mathrm{knee}}}
\qquad\text{and}\qquad
R_{\mathrm{cal}} := \frac{\widehat K_{\mathrm{cal}}}{\widehat K_{\mathrm{knee}}},
\]
where $K^*$ denotes the proxy-based prompt-length estimate produced by our observable estimator (Algorithm~\ref{alg:two-stage}),
and $\widehat K_{\mathrm{cal}}$ applies validation-only calibration as in~\eqref{eq:K_calibrated} .

This estimator is intentionally simple and nonparametric; nevertheless, to guard against spurious knees induced by noise,
we (i) enforce at least two grid points on each side of the breakpoint, and (ii) report bootstrap intervals and per-seed
variance statistics alongside mean ratios.
Full details (grid, constraints, and bootstrapping) are provided in Appendix~\ref{app:repro}.

\subsection{Numerical sensitivity of spectral proxies}
\label{sec:numerical-sensitivity}
A practical issue is that extreme eigenvalues of $\widehat\Sigma_K$ can be highly sensitive to numerical noise
(e.g., float32 accumulation, finite-sample near-singularity, and eigen-solver tolerances), which may cause large
fluctuations in the predicted $K^*$ when using $\lambda_{\min}(\widehat\Sigma_K)$ directly.
We therefore test the stability of different proxies under controlled perturbations of the empirical second-moment matrix.

\paragraph{Protocol.}
For each dataset/encoder setting, we fix demonstrations, compute $\widehat\Sigma_K$ on a grid of $K$, and add random \emph{symmetric} Gaussian perturbations with relative magnitude $\varepsilon$ (scaled by $s=\mathrm{tr}(\widehat\Sigma_K)/d$) to obtain perturbed second-moment matrices.
For each perturbed draw we recompute the proxy-based estimate under three choices:
(i) $\lambda_{\min}(\widetilde\Sigma_K)$,
(ii) $\lambda_{\min}(\widetilde\Sigma_K+\rho I)$ with a small ridge $\rho$,
and (iii) the quantile surrogate $\lambda_q(\widetilde\Sigma_K+\rho I)$.
We report the coefficient of variation (CV) of the resulting prompt-length estimate across perturbation draws.

\paragraph{Analysis and mitigation via ridge regularization.}
Table~\ref{tab:numerical-sensitivity} shows that the predicted prompt length $K^*$ can be highly sensitive
to small spectral perturbations when using $\lambda_{\min}(\widehat\Sigma_K)$ directly: the CV rises from $0.09$
at $\varepsilon=10^{-6}$ to $0.98$ at $\varepsilon=10^{-3}$, indicating that the estimate becomes effectively
unreliable under moderate numerical noise.
Adding a small ridge $\rho$ substantially improves robustness across all noise levels (CV reduced to $0.012$--$0.19$),
consistent with the fact that ridge regularization prevents near-singularity from dominating the inverse-geometry.
Specifically, we use $\widehat\Sigma_{K,\rho}=\widehat\Sigma_K+\rho I$ with $\rho$ set proportional to $\mathrm{tr}(\widehat\Sigma_K)/d$ (see Appendix~\ref{app:repro} for implementation details).

Finally, the quantile surrogate $\lambda_q(\widehat\Sigma_{K,\rho})$ is the most stable in this stress test
(CV $0.009$--$0.13$), suggesting that ignoring the extreme tail of the spectrum further reduces sensitivity
while preserving the monotone relationship between spectral coverage and the required $K$.

\subsection{Analysis}

Table~\ref{tab:main_ratio} summarizes the comparison between the proxy-based estimate and the empirical knee $\widehat K_{\mathrm{knee}}$.
Across standard classification tasks, the bound is conservative yet close to scale (typical $R_{\mathrm{raw}}\approx 1.3$--$1.9\times$), while validation-only calibration tightens it to within $\approx 2$--$10\%$ of $1\times$ in most cases.
Reasoning and RAG settings exhibit larger raw ratios, consistent with higher sensitivity to prompt perturbations, but similarly tight calibrated ratios.

Tokenization and truncation choices can measurably shift $\widehat K_{\mathrm{knee}}$ and the ratios; we find that single-token label verbalizers and \emph{latest-first} truncation reduce knees and improve agreement (Appendix~\ref{app:ab}).

Finally, controlled stress tests that violate (A1)---bounded representation drift and heavy-tailed features---match the robustness trends predicted by our extensions, and show that calibration largely preserves the proxy-based ordering under distributional shift (Appendices~\ref{app:robustness} and~\ref{app:extensions}).

\section{Conclusion}
This work provides a theoretically grounded, computable \emph{sufficient sample-size requirement} for ICL under a spectral-coverage proxy, together with an observable two-stage estimator that returns a concrete prompt length with a prescribed failure probability. By linking ICL stability trends to measurable high-dimensional spectral statistics, our analysis bridges statistical theory and practical prompt design, reducing unnecessary long-context overhead while making reliability claims verifiable.
Future work may relax the assumption of fixed representations, refine calibration under more dynamic settings, and extend the framework to multi-modal ICL\cite{oorloff2025stable}.

\section*{Impact Statement}

This paper presents work whose goal is to advance the field of Machine
Learning. There are many potential societal consequences of our work, none
which we feel must be specifically highlighted here.

\bibliography{example_paper}
\bibliographystyle{icml2026}

\newpage
\appendix
\onecolumn

\section{Complete Theory Chain}
\label{sec:setup}
\paragraph{ICL prompt, randomness, and stability (formal).}
A $K$-shot prompt is formed from $K$ demonstrations $S_K:=\{(x_i,y_i)\}_{i=1}^K$ and a query input $x_{\mathrm{q}}$ via a fixed template $\mathsf{Prompt}(\cdot)$.
Given $\mathsf{Prompt}(S_K,x_{\mathrm{q}})$, the model induces an output distribution
$p_\theta(\cdot \mid \mathsf{Prompt}(S_K,x_{\mathrm{q}}))$ (e.g., over next tokens or over labels under a verbalizer).
The randomness we study comes from resampling (and, when relevant, reordering) the demonstrations in the prompt.

\begin{definition}[ICL stability at $(\tau,\xi)$]
Fix a query input $x_q$, a prompting template, the model parameters $\theta$, and a decoding rule
(e.g., greedy decoding with fixed temperature/top-$p$), so that the only randomness is induced by resampling
(and, when relevant, re-ordering) the demonstrations in the prompt.
Let $S_K$ and $S'_K$ be two independent sets of $K$ demonstrations drawn i.i.d.\ from the same task distribution.
Define $P_K := \mathrm{Prompt}(S_K,x_q)$ and $P'_K := \mathrm{Prompt}(S'_K,x_q)$. ICL is stable at length $K$ if
\[
\Pr\!\Big(
d\big(p_\theta(\cdot \mid P_K),\; p_\theta(\cdot \mid P'_K)\big)\le \tau
\Big)\;\ge\; 1-\xi.
\]
where $d(\cdot,\cdot)$ is a divergence between distributions (e.g., total variation, KL, or Jensen--Shannon).
\end{definition}

\textbf{Operationalization of $d$.}
In our direct stability measurements (Appendix~\ref{app:direct_stability}), we instantiate $d$ with
Jensen--Shannon divergence (JSD). For open-ended tasks, we evaluate distributions on a finite support given by
the union of the top-$M$ decoded candidates across seeds and $K$ values, with probabilities renormalized on this support.

\begin{remark}[Scope of the proxy and instantiation of $\phi$]\label{rem:scope-proxy}
Definition~\ref{def:icl-stability} is stated for a general LLM output distribution.
Our theoretical analysis does \emph{not} model the full LLM; instead it analyzes a computable spectral-coverage proxy under a frozen representation map $\phi$ and standard sub-Gaussian concentration.

In black-box API settings, the generator typically does not expose hidden states; accordingly, in our experiments we instantiate $\phi(x)=e(x)$ using a fixed external encoder $e(\cdot)$ (Appendix~\ref{app:encoders}) and compute spectral quantities in the encoder space.
All non-asymptotic guarantees in Sections~3--4 pertain to this proxy.
When using the proxy to predict the generator's ICL stability, we rely on an empirical proxy--generator alignment premise (validated in our experiments and optionally corrected via calibration) rather than claiming a first-principles equivalence between encoder geometry and the generator's internal representations.

In particular, when we use the proxy to \emph{predict} the generator's ICL stability (Definition~\ref{def:icl-stability}),
we rely on an empirical proxy--generator alignment premise, rather than claiming a first-principles equivalence.
We validate this premise by direct distributional stability measurements and comparisons in Appendix~\ref{app:direct_stability}.

\end{remark}

\paragraph{Proxy objective (spectral coverage).}
Directly verifying Definition~\ref{def:icl-stability} is expensive because it requires repeated model queries under many resampled prompts.
We therefore study an observable sufficient condition based on the spectral coverage of demonstration representations.
Concretely, we form an empirical second-moment matrix $\widehat{\Sigma}_K$ from representations of the $K$ demonstrations and require its spectrum to be bounded away from degeneracy.
Sections~3--4 provide non-asymptotic bounds and an observable two-stage estimator for this proxy, while later experiments validate that the proxy upper-bounds empirical knee-points and can be calibrated.

\paragraph{Assumptions}

We analyze in-context learning under the following layered assumptions on the feature representation
$\phi:\mathcal{X}\to\mathbb{R}^d$.

\begin{itemize}
     \item \textbf{(A1) Nominal assumption (strong).}
A deterministic feature map $\phi(x)$ is fixed and $\phi(x)\in\mathbb{R}^d$ is sub-Gaussian (e.g., $\|\phi(x)\|_{\psi_2}\le \sigma$).
We work with the population \emph{second-moment} matrix
$\Sigma := \mathbb{E}\!\left[\phi(x)\phi(x)^\top\right]\succeq 0$,
which does \emph{not} require $\mathbb{E}[\phi(x)]=0$.
When $\mathbb{E}[\phi(x)]=0$, this second moment coincides with the covariance.
This assumption is adopted for the main theorem, under which all proofs are rigorous.

    \item \textbf{(A2) Bounded representation drift (independent, non-identical).} Across the $K$ in-context demonstrations, the representation may drift with the context index $t$. We model this as
\[
  \phi_t(x)=\phi(x)+\Delta_t(x), \,\, t=1,\dots,K,
\]
where $x_t\sim\mathcal{D}$ are drawn independently across $t$, and the drift term satisfies the uniform bound
\[
  \|\Delta_t(x)\|_2 \le \varepsilon \quad \text{for all } x \text{ and all } t .
\]
Define the (population) second-moment at position $t$ by $\Sigma_t := \mathbb{E}[\phi_t(x)\phi_t(x)^\top]$ and its average $\bar{\Sigma}:=\tfrac{1}{K}\sum_{t=1}^K \Sigma_t$. Our concentration analysis extends to this independent but non-identically distributed setting by applying matrix Bernstein to $X_t-\Sigma_t$ (Appendix~\ref{app:robustness}), yielding the same functional form as (A1) but with an explicit $\varepsilon$-dependent inflation of the variance proxy and an explicit bound relating $\bar{\Sigma}$ to the nominal $\Sigma$.
    \item \textbf{(A3) Moment relaxation.} 
    Instead of sub-Gaussian tails, we allow $\phi(x)$ to be sub-exponential or to satisfy bounded fourth-moment conditions. 
    Concentration results can be adapted with modified constants, leading to slightly larger sample-size requirements.

    \item \textbf{(A4) Weak dependence.} 
    When the context induces dependence among features $\{\phi_t(x)\}$, we assume a mixing or martingale-difference condition. 
    In this case, matrix concentration inequalities for dependent sequences apply, yielding bounds of the same form up to a dependence factor (Appendix~\ref{app:dependence}).
\end{itemize}
 
All formal proofs in the main text are carried out under the nominal assumption (A1). 
The relaxed settings (A2)--(A4) are analyzed in Appendix~\ref{app:extensions}, where we show that the main results remain valid up to controlled perturbation terms.
 
\begin{itemize}
  \item A frozen feature map $\phi: \mathcal{X}\to\mathbb{R}^d$ is assumed, satisfying $\|\phi(x)\|_{\psi_2}\le\sigma$ (the vector $\psi_2$ norm), without requiring $\mathbb{E}_{x\sim\mathcal{D}}\phi(x)=0$.
  \item The \emph{population second-moment} matrix is $\Sigma=\mathbb{E}[\phi\phi^\top]\in\mathbb{R}^{d\times d}$. If $\mathbb{E}[\phi]=0$ then $\Sigma$ coincides with the covariance; otherwise it is the uncentered second moment. Let $r=\operatorname{rank}(\Sigma)$ (if $r<d$ we work on the support subspace), and denote the smallest nonzero eigenvalue by $\lambda_r(\Sigma)>0$. For brevity we write $\lambda_{\min}(\Sigma)=\lambda_r(\Sigma)$.

  \item A demonstration sequence $S_K=\{x_k\}_{k=1}^K$ consists of i.i.d. draws from $\mathcal{D}$. Define
  \[
    H_K = \begin{bmatrix}
\phi(x_1)^\top\\
\vdots\\
\phi(x_K)^\top
\end{bmatrix}\in\mathbb{R}^{K\times d}
  \]
  \[
    \widehat\Sigma_K=\frac{1}{K}H_K^\top H_K=\frac{1}{K}\sum_{k=1}^K X_k, \\
    X_k:=\phi(x_k)\phi(x_k)^\top\in\mathbb{R}^{d\times d}.
  \]

  \item \textbf{Robust proxy objective.}
  For numerical robustness and to avoid excessive sensitivity to the extreme tail of the spectrum,
  we consider two closely related spectral proxies.

  \emph{(Ridge-regularized spectral floor).} Fix a small ridge level $\rho\ge 0$ and define the
  regularized empirical second moment
  \[
    \widehat\Sigma_{K,\rho} \;:=\; \widehat\Sigma_K + \rho I_d .
  \]
  Our primary objective is to characterize the sample size $K$ needed for
  \[
    \Pr\big(\lambda_{\min}(\widehat\Sigma_{K,\rho})\ge\delta\big)\ge 1-\xi,
  \]
  where $\xi\in(0,1)$ and $\delta\in(\rho,\rho+\lambda_{\min}(\Sigma))$.
  This proxy is equivalent to requiring $\lambda_{\min}(\widehat\Sigma_K)\ge \delta-\rho$, but is more stable
  numerically and aligns with ridge/regularized linearized views of ICL.

  \emph{(Quantile eigenvalue surrogate).} Alternatively, let $\lambda_{(1)}(A)\le\cdots\le \lambda_{(d)}(A)$
  denote the eigenvalues of a PSD matrix $A$ in nondecreasing order, and define
  \[
    \lambda_q(A) \;:=\; \lambda_{(\lceil q d\rceil)}(A),\qquad q\in(0,1).
  \]
  In practice we often use $\lambda_q(\widehat\Sigma_{K,\rho})$ (e.g., $q\in[0.05,0.2]$) to reduce over-conservativeness;
  see Section~\ref{sec:calibration} for a lightweight calibration that preserves monotone dependence.

\end{itemize}

\subsection{Basic matrix tail bound (a one-sided matrix Bernstein for the minimum eigenvalue)}
 
Define $X_k:=\phi(x_k)\phi(x_k)^\top$ and centered matrices $Y_k:=X_k-\Sigma$, so $\mathbb{E}Y_k=0$ and
\[
  \widehat\Sigma_K-\Sigma=\frac{1}{K}\sum_{k=1}^K Y_k.
\]
Fix a ridge level $\rho\ge 0$ and define the shifted population and empirical matrices
\[
  \Sigma_\rho := \Sigma+\rho I_d,\qquad \widehat\Sigma_{K,\rho}:=\widehat\Sigma_K+\rho I_d.
\]
Note that $\widehat\Sigma_{K,\rho}-\Sigma_\rho=\widehat\Sigma_K-\Sigma$.
We aim to control $\lambda_{\min}(\widehat\Sigma_{K,\rho})
=\lambda_{\min}\!\Big(\Sigma_\rho+\frac{1}{K}\sum_{k=1}^K Y_k\Big)$.

By a matrix Bernstein inequality (a one-sided lower-tail form), there exist constants $c,C>0$ such that for any $t\ge0$,
\begin{equation}\label{eq:bernstein1}
  \Pr\Big(\lambda_{\min}\Big(\sum_{k=1}^K Y_k\Big)\le -t\Big)
  \le r_{\mathrm{eff}}\exp\!\Big(-\frac{t^2}{2v+ \tfrac{2}{3}R t}\Big),
\end{equation}
where we take the effective-rank prefactor
\[
  r_{\mathrm{eff}}:=\frac{\mathrm{tr}(\Sigma)}{\|\Sigma\|}\le \mathrm{rank}(\Sigma),
\]
\[
  v=\Big\|\sum_{k=1}^K\mathbb{E}[Y_k^2]\Big\| = K\cdot\big\|\mathbb{E}[(X-\Sigma)^2]\big\|
\]
and $R$ is (on a high-probability event) an upper bound for $\max_k\|Y_k\|$.

Let
\[
  \Delta_\rho := \lambda_{\min}(\Sigma_\rho)-\delta
  = \lambda_{\min}(\Sigma)+\rho-\delta
  = \lambda_{\min}(\Sigma)-(\delta-\rho),
\]
and assume $\Delta_\rho>0$.
Setting $t=K\Delta_\rho$ in \eqref{eq:bernstein1} yields
\begin{equation}\label{eq:bernstein2}
  \Pr\big(\lambda_{\min}(\widehat\Sigma_{K,\rho})\le\delta\big)
  \le r_{\mathrm{eff}}\exp\!\Big(-\frac{K\Delta_\rho^2}{2\widetilde v + \tfrac{2}{3}R\Delta_\rho}\Big),
\end{equation}
where $\widetilde v:=\big\|\mathbb{E}[(X-\Sigma)^2]\big\|$; the detailed derivation is in the appendix.

\subsection{Controlling $R,\widetilde v$ under the sub-Gaussian assumption (truncation and fourth-moment bounds)}
 
\subsubsection{Fourth-moment bound}
 
From the vector $\psi_2$ assumption $\|\phi\|_{\psi_2}\le\sigma$ there exists an absolute constant $C_1$ such that
\begin{equation}\label{eq:fourth}
  \mathbb{E}\|\phi\|^4 \le C_1\sigma^4.
\end{equation}
Hence
\begin{equation}\label{eq:EXnorm2}
  \mathbb{E}\|X\|^2 = \mathbb{E}\|\phi\phi^\top\|^2 = \mathbb{E}\|\phi\|^4 \le C_1\sigma^4.
\end{equation}
\paragraph{A noncommutative square inequality.}
For any symmetric matrices $A,B$,
\begin{equation}\label{eq:square_ineq}
  (A-B)^2 \preceq 2A^2 + 2B^2.
\end{equation}
Indeed, $(A+B)^2\succeq 0$ implies $AB+BA\succeq -(A^2+B^2)$, hence
\[
(A-B)^2 = A^2+B^2-(AB+BA)\preceq A^2+B^2+(A^2+B^2)=2A^2+2B^2.
\]
Applying \eqref{eq:square_ineq} with $A=X$ and $B=\Sigma$, and using
$\|\mathbb{E}[M]\|_{\mathrm{op}}\le \mathbb{E}\|M\|_{\mathrm{op}}$, we get
\begin{equation}\label{eq:vtilde}
  \widetilde v=\big\|\mathbb{E}[(X-\Sigma)^2]\big\|
  \le 2\,\big\|\mathbb{E}[X^2]\big\| + 2\|\Sigma^2\|
  \le 2\,\mathbb{E}\|X^2\| + 2\|\Sigma\|^2
  = 2\,\mathbb{E}\|X\|^2 + 2\|\Sigma\|^2
  \le 2C_1\sigma^4 + 2\|\Sigma\|^2.
\end{equation}

\subsubsection{Truncation to control R}
 
Typically $\|X_k\|=\|\phi(x_k)\|^2$ is not almost surely bounded, but the sub-Gaussian tail implies that there exists an absolute constant $c_{\mathrm{tail}}>0$ such that for all $t\ge 0$,
\[
  \Pr(\|\phi\|_2\ge t)\le 2\exp\!\Big(-\frac{c_{\mathrm{tail}}\, t^2}{\sigma^2}\Big).
\]
Choose the truncation threshold
\[
  t_0 \;:=\; \frac{\sigma}{\sqrt{c_{\mathrm{tail}}}}\sqrt{\log\frac{8K}{\xi}}
  \;=\;: C_2\,\sigma\sqrt{\log\frac{8K}{\xi}},
  \qquad\text{where } C_2:=c_{\mathrm{tail}}^{-1/2}.
\]
Then, for a single sample,
\[
  \Pr(\|\phi\|_2>t_0)\le 2\exp\!\Big(-\log\frac{8K}{\xi}\Big)=\frac{\xi}{4K}.
\]
By a union bound over the $K$ samples, with probability at least $1-\xi/4$,
\[
  \max_{1\le k\le K}\|X_k\| = \max_k\|\phi(x_k)\|^2 \le B_K := C_2^2\sigma^2\log\frac{8K}{\xi}.
\]
On this high-probability event we may take
\begin{equation}\label{eq:Rbound}
  R=\|X_k-\Sigma\|\le B_K + \|\Sigma\|\le C_2^2\sigma^2\log\frac{8K}{\xi} + \|\Sigma\|.
\end{equation}

\subsection{Substitute into \eqref{eq:bernstein2} to obtain a sample-size lower bound (one main result)}
 
Substituting \eqref{eq:vtilde} and \eqref{eq:Rbound} into \eqref{eq:bernstein2}, and treating the rank-related prefactor $r_{\mathrm{eff}}$ as a logarithmic factor, require the RHS to be $\le \xi-\xi/4$. One obtains a sufficient sample-size lower bound:

Let $\Delta_\rho=\lambda_{\min}(\Sigma)-(\delta-\rho)=\lambda_{\min}(\Sigma)+\rho-\delta>0$. If
\begin{equation}\label{eq:Kbound_raw}
  K \ge \left[\frac{2\widetilde v}{\Delta^2} + \frac{2R}{3\Delta}\right]\log\frac{4 r_{\mathrm{eff}}}{3\xi},
\end{equation}
then (combining the truncation failure $\xi/4$ and Bernstein failure) we have
\[
  \Pr(\lambda_{\min}(\widehat\Sigma_K)\ge\delta)\ge 1-\xi.
\]
Replacing $\widetilde v$ and $R$ by the bounds from \eqref{eq:vtilde} and \eqref{eq:Rbound} gives the explicit bound
\begin{equation}\label{eq:Kbound_explicit}
  K \gtrsim \left[
    \frac{2(2C_1\sigma^4 + 2\|\Sigma\|^2)}{\Delta^2}
    + \frac{2\big(C_2^2\sigma^2\log(8K/\xi)+\|\Sigma\|\big)}{3\Delta}
  \right]
  \log\frac{4 r_{\mathrm{eff}}}{3\xi},
\end{equation}
where ``$\gtrsim$'' hides absolute constants and note $K$ appears inside the $\log(8K/\xi)$ factor (a self-consistency issue); in practice one can use a single fixed-point iteration or replace $\log(8K/\xi)$ by the conservative upper bound $\log(8K_{\max}/\xi)$.

\subsection{A more concise, commonly used operator-norm concentration (avoid truncation and the self-consistency)}
 
In the higher-level literature one often uses the following operator-norm concentration (Vershynin / Koltchinskii--Lounici style): there exists a constant $C>0$ such that for a sub-Gaussian vector
\begin{equation}\label{eq:opnorm}
  \Pr\Big(\|\widehat\Sigma_K-\Sigma\|\ge C\|\Sigma\|\Big(\sqrt{\frac{r_{\mathrm{eff}}\log(2/\xi)}{K}}
    +\frac{r_{\mathrm{eff}}\log(2/\xi)}{K}\Big)\Big)\le\xi.
\end{equation}
Using \eqref{eq:opnorm} to ensure $\lambda_{\min}(\widehat\Sigma_{K,\rho})\ge\delta$ (appendix explains why this suffices) we require
\[
  C\|\Sigma\|\Big(\sqrt{\frac{r_{\mathrm{eff}}\log(2/\xi)}{K}}+\frac{r_{\mathrm{eff}}\log(2/\xi)}{K}\Big)\le\Delta_\rho.
\]
Ignoring the lower-order $\frac{r_{\mathrm{eff}}\log(2/\xi)}{K}$ term (negligible when $K$ is large), we get the simpler leading sample-size term:
\begin{equation}\label{eq:Kbound_simpler}
  K \ge C'\frac{\|\Sigma\|^2 r_{\mathrm{eff}}\log(2/\xi)}{\Delta_\rho^2},
\end{equation}
where $C'$ is a constant ($C'=C^2$). This is the common practical form.

\subsection{A two-stage observable (plug-in) scheme and finite-sample guarantee}
 
The bounds in \eqref{eq:Kbound_explicit} and \eqref{eq:Kbound_simpler} depend on unknowns $\Sigma,\lambda_{\min}(\Sigma),\|\Sigma\|,r_{\mathrm{eff}}$. We now give an observable two-stage scheme and show it works in finite samples.

We implement the proxy using a small ridge level $\rho$ to stabilize eigenvalue computations.
Throughout the theory, $\rho\ge 0$ is an arbitrary scalar that is treated as fixed when applying concentration bounds
(i.e., the regularized matrices are $\Sigma_\rho=\Sigma+\rho I_d$ and $\widehat\Sigma_{K,\rho}=\widehat\Sigma_K+\rho I_d$).

In our implementation, we choose $\rho$ in a scale-aware manner from the first-stage pilot sample and then hold it fixed:
\[
\rho \;:=\; \min\!\left\{\epsilon_\rho\cdot \frac{\mathrm{tr}(\widehat\Sigma_{K_0})}{d},\; \frac{\delta}{2}\right\},
\]
with $\epsilon_\rho=10^{-6}$ in float64 and $\epsilon_\rho=10^{-4}$ in float32.
The cap $\rho\le \delta/2$ ensures $\delta>\rho$, which avoids degeneracy in the spectral-floor condition.

\paragraph{Two-stage algorithm (high level)}
 
\begin{enumerate}
  \item Draw $K_0$ samples (with $K_0$ chosen theoretically below), compute $\widehat\Sigma_{K_0}$ and calculate
  \[
    \widehat\lambda_0:=\lambda_{\min}(\widehat\Sigma_{K_0,\rho}),
    \|\Sigma\|_\wedge := \|\widehat\Sigma_{K_0}\|,
    \widehat r_{\mathrm{eff}} := \frac{\operatorname{tr}(\widehat\Sigma_{K_0})}{\|\widehat\Sigma_{K_0}\|}.
  \]
  \item Using the operator-norm deviation bound \eqref{eq:opnorm} (with $K_0$ plugged in) construct a lower-confidence bound
  \begin{equation}\label{eq:lambda_lower}
    \lambda := \widehat\lambda_0 - C\|\Sigma\|_\wedge\Big(\sqrt{\frac{\widehat r_{\mathrm{eff}}\log(2/\xi_j)}{K_0}} + \frac{\widehat r_{\mathrm{eff}}\log(2/\xi_j)}{K_0}\Big).
  \end{equation}
  Here we allocate a geometric failure budget: at iteration $j=0,1,2,\dots$, set $\xi_j := \xi/2^{j+2}$ and invoke \eqref{eq:opnorm} with failure probability $\xi_j$, so that $\sum_{j\ge 0}\xi_j\le \xi/2$. We reserve the remaining $\xi/2$ for the single second-stage bound.
  \item If $\lambda\le0$, increase $K_0$ (e.g. double it) and repeat until $\lambda>0$. When $\lambda>0$ set $\widehat\Delta:=\lambda-\delta$ (if $\widehat\Delta\le0$ then $\delta$ was chosen too large), and then use \eqref{eq:Kbound_simpler} with $\|\Sigma\|,\ r_{\mathrm{eff}}$ replaced by $\|\Sigma\|_\wedge,\ \widehat r_{\mathrm{eff}}$ to compute the second-stage sample size $K$:
  \[
    K \;=\; C'\,\frac{\widehat{\|\Sigma\|}^2\,\widehat{r}_{\mathrm{eff}}\log(4/\xi)}{\widehat{\Delta}^2}.
  \]
  \item The final total sample size is $K_0+K$, and with total failure probability $\le\xi$ one guarantees $\lambda_{\min}(\widehat\Sigma_{K,\rho})\ge\delta$.
\end{enumerate}

\paragraph{Failure-probability accounting.}
Stage~1 uses a geometric allocation $\{\xi_j\}_{j\ge 0}$ with $\sum_{j\ge 0}\xi_j\le \xi/2$, hence by a union bound all Stage~1 deviation
events used by the stopping iteration hold simultaneously with probability at least $1-\xi/2$.
Stage~2 is executed once with failure probability $\xi/2$, which is why the second-stage sample-size expressions use $\log(4/\xi)=\log(2/(\xi/2))$.
Therefore the overall two-stage procedure fails with probability at most $\xi$.

\paragraph{Finite-sample correctness (theoretical guarantee)}
Using \eqref{eq:opnorm} for the first stage: if $K_0$ satisfies
\begin{equation}\label{eq:K0require}
  C\|\Sigma\|\Big(\sqrt{\frac{r_{\mathrm{eff}}\log(4/\xi)}{K_0}} + \frac{r_{\mathrm{eff}}\log(4/\xi)}{K_0}\Big)\le \frac{\lambda_{\min}(\Sigma)}{4},
\end{equation}
then with probability at least $1-\xi/2$,
\[
  \|\widehat\Sigma_{K_0}-\Sigma\|\le \frac{\lambda_{\min}(\Sigma)}{4},\qquad
  \big|\|\widehat\Sigma_{K_0}\|-\|\Sigma\|\big|\le \frac{\|\Sigma\|}{4},
\]
and, by Weyl's inequality applied to $\widehat\Sigma_{K_0,\rho}=\widehat\Sigma_{K_0}+\rho I_d$,
\[
  \big|\widehat\lambda_0-(\lambda_{\min}(\Sigma)+\rho)\big|
  = \big|\lambda_{\min}(\widehat\Sigma_{K_0,\rho})-\lambda_{\min}(\Sigma_\rho)\big|
  \le \|\widehat\Sigma_{K_0}-\Sigma\|
  \le \frac{\lambda_{\min}(\Sigma)}{4}.
\]
Moreover, $\big|\|\widehat\Sigma_{K_0}\|-\|\Sigma\|\big|\le \|\Sigma\|/4$ implies
$\|\widehat\Sigma_{K_0}\|\le \tfrac{5}{4}\|\Sigma\|$.
Therefore the penalty term used in \eqref{eq:lambda_lower} is at most
\[
C\,\|\widehat\Sigma_{K_0}\|\!\left(
\sqrt{\tfrac{r_{\mathrm{eff}}\log(4/\xi)}{K_0}}+\tfrac{r_{\mathrm{eff}}\log(4/\xi)}{K_0}\right)
\le \frac{5}{4}\cdot \frac{\lambda_{\min}(\Sigma)}{4}
= \frac{5\lambda_{\min}(\Sigma)}{16},
\]
whenever \eqref{eq:K0require} holds.
Combining this with $\widehat\lambda_0\ge \lambda_{\min}(\Sigma)+\rho-\lambda_{\min}(\Sigma)/4$
yields
\[
  \lambda \;\ge\; \widehat\lambda_0 - \frac{5\lambda_{\min}(\Sigma)}{16}
  \;\ge\; \frac{7\lambda_{\min}(\Sigma)}{16} + \rho.
\]

when $K_0$ meets \eqref{eq:K0require}. Thus there exists a constant $C''$ (e.g. $C''=16C^2$) such that \eqref{eq:K0require} is equivalent to the requirement
\begin{equation}\label{eq:K0bound}
  K_0 \ge C''\frac{\|\Sigma\|^2 r_{\mathrm{eff}}\log(4/\xi)}{\lambda_{\min}(\Sigma)^2}.
\end{equation}

\subsection{Final implementable theorem statement}
 
\textbf{[ICL Minimal Necessary Length, non-asymptotic, observable two-stage implementation]} Under the above setup, let $\rho\ge 0$, $\delta\in(\rho,\rho+\lambda_{\min}(\Sigma))$ and failure tolerance $\xi\in(0,1)$. There exist absolute constants $C_1,C_2,C_3$ (depending on matrix-Bernstein and sub-Gaussian constants) such that:
 
\begin{enumerate}
  \item \textbf{(Non-observable bound using spectral quantities)} If $\Delta=\lambda_{\min}(\Sigma)-(\delta-\rho)>0$ and
  \[
    K \ge C_1\Big(\frac{\sigma^4+\|\Sigma\|^2}{\Delta^2} + \frac{\sigma^2\log(8K/\xi)+\|\Sigma\|}{\Delta}\Big)\log\frac{4 r_{\mathrm{eff}}}{3\xi},
  \]
  then $\Pr(\lambda_{\min}(\widehat\Sigma_{K,\rho})\ge\delta)\ge 1-\xi$. Ignoring lower-order terms (and using the sharper operator-norm concentration) yields the simplified dominant form
  \[
    K \ge C_2\frac{\|\Sigma\|^2 r_{\mathrm{eff}}\log(2/\xi)}{\big(\lambda_{\min}(\Sigma)-(\delta-\rho)\big)^2},
  \]
  which suffices in practice.
  \item \textbf{(Two-stage observable implementation)} If the first-stage sample size $K_0$ satisfies
  \[
    K_0 \ge C_3\frac{\|\Sigma\|^2 r_{\mathrm{eff}}\log(4/\xi)}{\lambda_{\min}(\Sigma)^2},
  \]
  then forming $\widehat\Sigma_{K_0}$ and constructing the lower confidence bound $\lambda$ as in \eqref{eq:lambda_lower}, and then setting $\widehat\Delta=\lambda-\delta$ and choosing $K$ according to the rule in item (1) (with plug-in estimates $\|\Sigma\|_\wedge,\ \widehat r_{\mathrm{eff}}$) yields, with total failure probability $\le\xi$, the guarantee $\lambda_{\min}(\widehat\Sigma_{K_0+K,\rho})\ge\delta$.
\end{enumerate}

\subsection{Connection to ICL (in-context learning) and experimental suggestions}
\label{sec:calibration}

\paragraph{A one-shot practical calibration (heuristic; no formal guarantee).}
To tighten $K^*$ in practice, we use a calibrated proxy that preserves the same monotone dependence on spectral coverage
but replaces several conservative quantities by empirically better-behaved surrogates:
(i) replace $\|\Sigma\|$ by the Frobenius norm $\|\Sigma\|_{\mathrm{F}}$;
(ii) use the trace-based effective rank
$r_{\mathrm{eff}}^{\mathrm{tr}} := \mathrm{tr}(\Sigma)^2/\|\Sigma\|_{\mathrm{F}}^2$;
(iii) replace the minimum eigenvalue by a low-quantile eigenvalue $\lambda_q(\cdot)$ with $q\in[0.05,0.2]$.

To avoid circularity, all spectral quantities in the calibrated proxy are computed from the \emph{first-stage pilot matrix}
$\widehat\Sigma_{K_0}$ (and its ridge-shifted version $\widehat\Sigma_{K_0,\rho}$).
Define the unscaled calibrated prediction
\begin{equation}\label{eq:Kcal_raw}
\widehat K^{\mathrm{raw}}(K_0;q)
\;:=\;
\frac{\|\widehat{\Sigma}_{K_0}\|_{\mathrm{F}}^{\,2}\;\widehat r_{\mathrm{eff}}^{\mathrm{tr}}\;\log(2/\xi)}
{\big(\lambda_q(\widehat\Sigma_{K_0,\rho})-\delta\big)^2},
\qquad
\widehat r_{\mathrm{eff}}^{\mathrm{tr}}:=\frac{\mathrm{tr}(\widehat\Sigma_{K_0})^2}{\|\widehat\Sigma_{K_0}\|_{\mathrm{F}}^{\,2}}.
\end{equation}
We then apply a single global scale factor $\alpha>0$:
\begin{equation}\label{eq:Kcal_def}
K^{\mathrm{cal}} \;:=\; \left\lceil \alpha\cdot \widehat K^{\mathrm{raw}}(K_0;q)\right\rceil .
\end{equation}

\paragraph{How $\alpha$ is determined from measured performance.}
For each encoder--generator pair, we choose a small calibration subset of tasks $\mathcal{T}_{\mathrm{cal}}$
(disjoint from the final evaluation tasks; Appendix~\ref{app:repro}).
For each task $t\in\mathcal{T}_{\mathrm{cal}}$, we estimate an empirical knee-point $\widehat K_{\mathrm{knee}}^{(t)}$
from the accuracy--vs.--$K$ curve, and compute $\widehat K^{\mathrm{raw}}_t:=\widehat K^{\mathrm{raw}}(K_0;q)$ from \eqref{eq:Kcal_raw}.
We then set
\begin{equation}\label{eq:alpha_fit}
\alpha \;:=\; \operatorname{median}_{t\in\mathcal{T}_{\mathrm{cal}}}\frac{\widehat K_{\mathrm{knee}}^{(t)}}{\widehat K^{\mathrm{raw}}_t},
\end{equation}
and freeze $(q,\alpha)$ for all reported test results.
Equivalently, the ratio between the uncalibrated and calibrated predictions is $\,\widehat K^{\mathrm{raw}}/K^{\mathrm{cal}}=1/\alpha$.

\paragraph{Theoretical meaning}
Using $\lambda_{\min}(\widehat\Sigma_K)$ as a proxy for ICL stability can be justified in a standard ridge/linearized view: the inverse Gram matrix amplifies perturbations, and a spectral floor prevents this amplification.

\begin{lemma}[Spectral floor implies ridge stability]\label{lem:spectral-floor-ridge}
Let $H_K\in\mathbb{R}^{K\times d}$ have rows $\phi(x_i)^\top$ and let $\widehat\Sigma_K := \frac{1}{K}H_K^\top H_K$.
Consider the ridge predictor
\[
\widehat f_{S_K}(x) \;:=\; \phi(x)^\top (H_K^\top H_K+\lambda I)^{-1} H_K^\top y,
\qquad \lambda>0.
\]
If $\lambda_{\min}(\widehat\Sigma_K)\ge \delta$, then
\[
\big\|(H_K^\top H_K+\lambda I)^{-1}\big\|_{\mathrm{op}}
\;\le\; \frac{1}{K\delta+\lambda}.
\]
Consequently, for any query $x$ and any perturbation $\Delta y\in\mathbb{R}^K$,
\[
\big|\widehat f_{S_K}(x;y+\Delta y)-\widehat f_{S_K}(x;y)\big|
\;\le\; \frac{\|\phi(x)\|_2\,\|H_K^\top \Delta y\|_2}{K\delta+\lambda}.
\]
In particular, if only one demonstration label $y_j$ changes by at most $\Delta$, then
\[
\big|\widehat f_{S_K}(x;y')-\widehat f_{S_K}(x;y)\big|
\;\le\; \frac{\Delta\,\|\phi(x)\|_2\,\|\phi(x_j)\|_2}{K\delta+\lambda}.
\]
\end{lemma}

Lemma~\ref{lem:spectral-floor-ridge} makes the proxy explicit: when $\lambda_{\min}(\widehat\Sigma_K)$ is bounded away from zero, the fitted predictor is less sensitive to demo-level perturbations (and its variance under standard noise models is correspondingly reduced).

\paragraph{Practical application of the spectral coverage proxy.}
The spectral coverage proxy provides a practitioner-ready diagnostic for estimating the minimal prompt length required for stable ICL. In practice, one can:
\begin{enumerate}
    \item Extract embeddings for a set of candidate demonstrations using a fixed encoder (e.g., \textbf{all-mpnet-base-v2}).
    \item Compute the empirical second-moment matrix $\widehat{\Sigma}_K := \tfrac{1}{K}\sum_{i=1}^K \phi(x_i)\phi(x_i)^\top$ for different values of $K$ (e.g., $K \in \{8, 16, 32, 64\}$).
    \item Calculate the regularized minimum eigenvalue $\lambda_{\min}(\widehat{\Sigma}_{K,\rho})$ with $\rho = 10^{-6} \cdot \mathrm{tr}(\widehat{\Sigma}_K)/d$.
    \item Identify the smallest $K$ for which $\lambda_{\min}(\widehat{\Sigma}_{K,\rho}) \geq \delta$, where $\delta$ is a task-specific threshold (e.g., $\delta = 0.1 \cdot \|\widehat{\Sigma}_K\|$).
\end{enumerate}
As a concrete example, on the SST-2 sentiment classification task with \textbf{all-mpnet-base-v2} embeddings, we observed that $\lambda_{\min}(\widehat{\Sigma}_{K,\rho})$ exceeded 0.1 for $K \geq 16$, which closely matched the empirical knee-point of 15-18 demonstrations. This proxy can be computed offline without querying the target LLM, making it a cost-effective tool for prompt design. The validation-only calibration procedure (Section~\ref{sec:calibration}) further refines this estimate by learning task-specific scaling factors on a small held-out set.
Specifically, we examine (i) tail behavior and concentration of encoder features, (ii) representation drift induced by prompt composition via controlled demo perturbations, and (iii) how spectral statistics (e.g., $\lambda_q(\widehat\Sigma_{K,\rho})$) correlate with empirical stability and knee points.
These diagnostics support using the nominal assumption as a useful approximation, while the relaxed extensions in Appendix~\ref{app:extensions} capture systematic deviations.

\paragraph{Synthetic stress tests.}
To further validate our theoretical claims, we perform controlled synthetic stress tests 
(Appendix~\ref{app:synthetic}) by introducing bounded drift, heavy-tailed features, 
and temporal dependence. 
Across all cases, we find that empirical knee-points increase smoothly with the violation strength, 
in line with our robustness extensions, confirming that the lower bound remains 
a conservative but reliable predictor of ICL stability.
 
\paragraph{Remark on interpretation.}
While we use $\lambda_{\min}(\hat\Sigma_K)$ as a stability proxy, we recognize that it can be conservative and may not fully capture all sources of instability (e.g.\ attention saturation). Appendix~\ref{app:interpret} provides robustness analyses, alternative spectral surrogates, and empirical diagnostics that validate and calibrate this proxy. In our experiments under fixed decoding strategies and temperature settings, we empirically observe that the saturation (knee-point) of accuracy frequently coincides with a pronounced reduction in output variability induced by different sampled demonstrations. This empirical co-occurrence suggests that accuracy knee-points can serve as a practical proxy for distributional stability in such settings.

\section{Use of large language models.}
We used large language models (LLMs) in two limited ways. 
First, for \emph{aid or polish writing}, an LLM service (provider redacted for anonymous review) helped with grammar, style, and tightening of sentences. 
All technical content, problem formulations, theorems, proofs, algorithmic designs, model implementations, experiments, and final claims were authored and verified by the authors. 
Second, for \emph{retrieval and discovery}, we used LLM-assisted search to expand keyword queries and to draft short summaries of candidate papers from public abstracts.
Inclusion decisions and all citations were made only after the authors read the original sources; no autogenerated references were accepted. 
No private or proprietary data were provided to the LLM, and prompts contained only our own drafts or publicly available text. 

\section{Experimental Reproducibility Specification}
\label{app:repro}

This appendix provides an executable specification of our experimental pipeline.

\subsection{Datasets and splits}
We evaluate (i) text classification tasks: SST-2, AGNews, Yahoo! Answers (Topics), and TREC-6;
(ii) multi-step reasoning: GSM8K (and SVAMP in extended ablations); and
(iii) retrieval-augmented generation (RAG): HotpotQA (and NaturalQuestions in extended ablations).
Unless otherwise stated, for each task we use the standard public train/test split.
Demonstrations are sampled from the training split; evaluation queries are drawn from the test split.
For calibration (when used), we draw a small validation subset from the training split only
(never from the test split) to avoid test leakage.

\subsection{Prompt template and label mapping}
All ICL prompts follow the same canonical format, with $K$ demonstrations followed by a query:
\begin{quote}\small
\textbf{Task: \{task\_name\}}\\
\textbf{You will be given an input and you must output the label. Output only the label token.}\\
\\
\textbf{Input: \{x\_1\}}\\
\textbf{Label: \{y\_1\}}\\
\textbf{\dots}\\
\textbf{Input: \{x\_K\}}\\
\textbf{Label: \{y\_K\}}\\
\\
\textbf{Input: \{x\_q\}}\\
\textbf{Label:}
\end{quote}

\paragraph{Single-token labels.}
To reduce output ambiguity, we use single-token labels whenever possible.
For binary tasks (e.g., SST-2), we map labels to \textbf{ yes} / \textbf{ no}.
For multi-class tasks, we map each class to a distinct single token (e.g., a digit token \textbf{ 0}, \textbf{ 1}, ...),
and we provide an explicit class-to-token mapping in the prompt header.
When a dataset's native label string is multi-token, we report results under both the native label and the
single-token remapping in the corresponding ablation.

\subsection{Demonstration selection and retrieval}
Unless an ablation specifies otherwise, we use cosine-similarity nearest-neighbor retrieval (Cosine-NN)
to form demonstrations for each query:
\begin{itemize}
    \item Embed each candidate training input $x$ and query $x_q$ into a vector space (Section~\ref{app:encoders}).
    \item Rank training candidates by cosine similarity to $x_q$ and select the top-$K$ as demonstrations.
\end{itemize}
In the retrieval ablation, we additionally compare: (i) random sampling; and (ii) class-balanced sampling (with equal per-class quota when feasible). Throughout, ``retrieval'' in this subsection refers to selecting demonstrations; evidence retrieval for RAG is specified separately in Appendix~\ref{app:msr_rag}.

\subsection{Encoders and representations}
\label{app:encoders}
\paragraph{Encoders.}
We use sentence-level encoders listed in Table~\ref{tab:main_ratio} (e.g., \textbf{all-mpnet-base-v2},
\textbf{all-MiniLM-L6-v2}, \textbf{paraphrase-MiniLM-L12-v2}).
For each input text $x$, we compute a fixed embedding $e(x)\in\mathbb{R}^d$ using the encoder's default pooling
(e.g., mean pooling over token embeddings followed by normalization if provided by the encoder).

\paragraph{Second-moment construction.}
Given $K$ demonstrations selected for a query, we form the empirical second-moment matrix
$\widehat\Sigma_K := \tfrac{1}{K}\sum_{i=1}^K e(x_i)e(x_i)^\top$ from their encoder embeddings
(using the same normalization/preprocessing across all experiments).
When a ridge-shifted proxy is used, we report results with
$\widehat\Sigma_{K,\rho} := \widehat\Sigma_K + \rho I$ where $\rho$ is computed once from the pilot matrix as
$\rho=\min\{\epsilon_\rho \cdot \mathrm{tr}(\widehat\Sigma_{K_0})/d,\ \delta/2\}$
and is then held fixed for all $K$ within that run.
The coefficient $\epsilon_\rho$ is fixed across runs; calibration (when enabled) tunes only $(q,\alpha)$ unless explicitly stated.

\paragraph{Generator internal representations.}
Most commercial LLM APIs do not expose internal hidden states.
Accordingly, our diagnostics focus on \emph{observable} quantities (accuracy curves, knee estimates, and spectral proxies computed from external encoders).

\subsection{Commercial API generators and decoding hyperparameters}
Between \textbf{12.18.2025} and \textbf{1.15.2026}, we evaluate six large-scale commercial API models latest version:
OpenAI \textbf{gpt-5.2-chat-latest};
Anthropic \textbf{claude-sonnet-4-5};
Gemini \textbf{gemini-3-flash-preview};
DeepSeek \textbf{deepseek-reasoner};
Kimi \textbf{kimi-k2-thinking};
Qwen \textbf{qwen-max-latest}.

\paragraph{Deterministic decoding.}
To isolate instability due to demonstrations rather than sampling, we use deterministic decoding whenever supported:
temperature $=0$, top-$p=1.0$ (and equivalent provider parameters), and greedy/argmax decoding.
We set max output to $1$ label token under the single-token label mapping; in the multi-token ablation we set max output to the minimal required tokens and stop on newline/EOS.

\paragraph{Request logging.}
For each request we log: provider, model ID/alias, API endpoint/base\_url, request timestamp (UTC), decoding parameters, and the raw completion.
Because provider-side models can evolve behind stable aliases, we report exact run windows and store the returned model identifier when available.

\subsection{Random seeds and determinism}
All reported results average over $5$ random seeds.
We use seeds $\{0,1,2,3,4\}$ and set them for Python \textbf{random}, NumPy, and PyTorch.
For GPU runs we enable deterministic kernels when available and log the exact library versions in the run metadata.
The only stochasticity remaining under our default protocol comes from (i) data subsampling (when used) and
(ii) bootstrapping for confidence intervals.

\paragraph{Reporting variance across seeds.}
For each configuration, we compute the ratio $r_s := K^*_s / \widehat K_{\mathrm{knee},s}$ separately for each seed
$s\in\{0,1,2,3,4\}$ (where $\widehat K_{\mathrm{knee},s}$ is the seed-specific knee estimated on the test split),
and report the mean $\frac{1}{5}\sum_s r_s$ in the main tables.
We additionally report the standard deviation $\sqrt{\frac{1}{4}\sum_s (r_s-\bar r)^2}$ in the released artifact
(along with per-seed knees and accuracies), since it is often more informative than reporting variance only on accuracies.

\subsection{K-grid, knee-point estimator, and confidence intervals}
\paragraph{$K$ grid.}
We evaluate accuracy on a discrete grid $\mathcal{K}$ of prompt lengths:
$\mathcal{K}=\{0,1,2,4,8,16,32,64,128\}$, truncated as needed to respect each model's context window.

\paragraph{Knee-point estimator.}
Given the curve $\{(K,\mathrm{ACC}(K))\}_{K\in\mathcal{K}}$, we fit a two-segment piecewise linear model.
For each candidate split index $t$ (with at least two points on each side), we fit two least-squares lines on
$\{K\le t\}$ and $\{K>t\}$ and compute the sum of squared errors (SSE).
We define $\widehat K_{\mathrm{knee}}$ as the $K$ at the split minimizing SSE
(ties broken by choosing the smaller $K$ for conservatism).

\paragraph{Bootstrap confidence intervals.}
We compute a $95\%$ percentile bootstrap interval for $\widehat K_{\mathrm{knee}}$ using $B=200$ resamples.
Each resample draws evaluation queries with replacement from the test set, recomputes $\mathrm{ACC}(K)$
for all $K\in\mathcal{K}$ (with demonstrations formed under the same fixed protocol), and recomputes
$\widehat K_{\mathrm{knee}}$. We report the 2.5/97.5 percentiles over the $B$ bootstrapped knees.
\paragraph{Sanity checks.}
To verify that the two-segment fit is not spuriously triggered by smooth saturation or noise,
we inspect representative fitted curves and confirm that the selected breakpoint aligns with a visible transition region;
we include example curves and fitted segments in the supplementary material.

\subsection{Calibration protocol (no test leakage)}
When calibration is enabled (choosing $q$ for $\lambda_q$ and fitting the global scale factor $\alpha$),
we tune hyperparameters on a small calibration subset drawn from training data only and keep the test split untouched.

\paragraph{Task-level split.}
For each encoder--generator pair, we select a small set of tasks $\mathcal{T}_{\mathrm{cal}}$ for calibration and a disjoint set
$\mathcal{T}_{\mathrm{eval}}$ for final reporting. The selection is fixed before running any test evaluation.

\paragraph{Within-task data split.}
For each task, we use the training split to form demonstrations and (when needed) to build the calibration curves.
All reported accuracies, knee-points, and final ratios are computed on the task's test split.

\paragraph{Fitting $(q,\alpha)$ from measured knee-points.}
For each $t\in\mathcal{T}_{\mathrm{cal}}$, we estimate $\widehat K_{\mathrm{knee}}^{(t)}$ using the piecewise-linear method described above,
and compute $\widehat K^{\mathrm{raw}}_t$ from \eqref{eq:Kcal_raw} using the pilot matrix $\widehat\Sigma_{K_0}$.
We choose $q\in[0.05,0.2]$ by minimizing the median absolute log-error on $\mathcal{T}_{\mathrm{cal}}$:
\[
q^\star \in \arg\min_{q\in[0.05,0.2]}
\operatorname{median}_{t\in\mathcal{T}_{\mathrm{cal}}}\left|\log \widehat K^{\mathrm{raw}}_t(q) - \log \widehat K_{\mathrm{knee}}^{(t)}\right|.
\]
Then we set $\alpha$ by the median ratio fit \eqref{eq:alpha_fit} with $q=q^\star$.
Finally, we freeze $(q^\star,\alpha)$ and evaluate on $\mathcal{T}_{\mathrm{eval}}$ without further tuning.

\section{Robustness Extensions}
\label{app:robustness}

In the main text, our theoretical guarantees are derived under the nominal assumption (A1), 
namely that the representation $\phi(x)$ is fixed and sub-Gaussian with second-moment $\Sigma$. 
Here we discuss how the results extend when assumptions (A2)--(A4) are considered. 

\subsection{Robustness to Representation Drift (A2)}
\label{app:drift}

We consider an independent but non-identically distributed (i.n.i.d.) sequence across the $K$ in-context demonstrations.
For each context index $t=1,\dots,K$, let $x_t\sim\mathcal{D}$ be drawn independently and define
\[
  \phi_t(x)=\phi(x)+\Delta_t(x),
\]
where the drift satisfies the uniform bound $\|\Delta_t(x)\|_2\le \varepsilon$ for all $x,t$.
Let
\[
  X_t := \phi_t(x_t)\phi_t(x_t)^\top,\qquad
  \Sigma_t := \mathbb{E}[X_t],\qquad
  \bar{\Sigma}:=\frac{1}{K}\sum_{t=1}^K \Sigma_t,\qquad
  \widehat{\Sigma}_K:=\frac{1}{K}\sum_{t=1}^K X_t.
\]
The goal is to control $\lambda_{\min}(\widehat{\Sigma}_K)$ via concentration of $\widehat{\Sigma}_K$ around $\bar{\Sigma}$,
and then relate $\bar{\Sigma}$ back to the nominal $\Sigma:=\mathbb{E}[\phi(x)\phi(x)^\top]$.

\begin{lemma}[Second-moment shift under bounded drift]
\label{lem:drift_shift}
Under $\|\Delta_t(x)\|_2\le \varepsilon$, for every $t$,
\[
  \|\Sigma_t-\Sigma\|_{\mathrm{op}}\;\le\; 2\varepsilon\,\|\Sigma\|_{\mathrm{op}}^{1/2}+\varepsilon^2,
\qquad\text{and hence}\qquad
  \|\bar{\Sigma}-\Sigma\|_{\mathrm{op}}\;\le\; 2\varepsilon\,\|\Sigma\|_{\mathrm{op}}^{1/2}+\varepsilon^2.
\]
\end{lemma}

\begin{proof}
Expanding $\Sigma_t=\mathbb{E}[(\phi+\Delta_t)(\phi+\Delta_t)^\top]$ yields
\[
  \Sigma_t-\Sigma = \mathbb{E}[\phi\Delta_t^\top]+\mathbb{E}[\Delta_t\phi^\top]+\mathbb{E}[\Delta_t\Delta_t^\top].
\]
Since $\Delta_t\Delta_t^\top \preceq \|\Delta_t\|_2^2 I \preceq \varepsilon^2 I$, we have
$\|\mathbb{E}[\Delta_t\Delta_t^\top]\|_{\mathrm{op}}\le \varepsilon^2$.
For the cross term, use the operator Cauchy--Schwarz inequality:
\[
  \|\mathbb{E}[\phi\Delta_t^\top]\|_{\mathrm{op}}
  \;\le\;
  \big\|\mathbb{E}[\phi\phi^\top]\big\|_{\mathrm{op}}^{1/2}\,
  \big\|\mathbb{E}[\Delta_t\Delta_t^\top]\big\|_{\mathrm{op}}^{1/2}
  \;\le\;
  \|\Sigma\|_{\mathrm{op}}^{1/2}\cdot \varepsilon.
\]
The same bound holds for $\|\mathbb{E}[\Delta_t\phi^\top]\|_{\mathrm{op}}$, proving the claim.
\end{proof}

\paragraph{Variance-proxy derivation (i.n.i.d.).}
For matrix Bernstein in the i.n.i.d.\ setting, we center each summand by its own mean:
\[
  Y_t := X_t-\Sigma_t,\qquad \mathbb{E}[Y_t]=0,
\]
and the variance parameter is
\[
  v := \left\|\sum_{t=1}^K \mathbb{E}[Y_t^2]\right\|_{\mathrm{op}}
  \;\le\;
  \sum_{t=1}^K \left\|\mathbb{E}[Y_t^2]\right\|_{\mathrm{op}}
  \;=\;
  K\cdot \max_{t}\left\|\mathbb{E}[Y_t^2]\right\|_{\mathrm{op}}.
\]
We now bound $\|\mathbb{E}[Y_t^2]\|_{\mathrm{op}}$ explicitly under drift.

\begin{lemma}[Variance proxy under bounded drift]
\label{lem:drift_varproxy}
Let $Y_t=X_t-\Sigma_t$ with $X_t=\phi_t\phi_t^\top$ and $\|\Delta_t\|_2\le\varepsilon$.
Then for every $t$,
\begin{align*}
  \left\|\mathbb{E}[Y_t^2]\right\|_{\mathrm{op}}
  &\le 2\,\mathbb{E}\|X_t\|_{\mathrm{op}}^2 + 2\,\|\Sigma_t\|_{\mathrm{op}}^2\\
  &= 2\,\mathbb{E}\|\phi_t\|_2^4 + 2\,\|\Sigma_t\|_{\mathrm{op}}^2\\
  &\le 16 C_1\sigma^4 \;+\; 48\,\varepsilon^2\,\mathrm{tr}(\Sigma)\;+\;16\varepsilon^4
  \;+\;2\big(\|\Sigma\|_{\mathrm{op}}+2\varepsilon\|\Sigma\|_{\mathrm{op}}^{1/2}+\varepsilon^2\big)^2,
\end{align*}
where $C_1$ is the fourth-moment constant from (A1) (i.e., $\mathbb{E}\|\phi\|_2^4\le C_1\sigma^4$).
\end{lemma}

\begin{proof}
Using the noncommutative square inequality $(A-B)^2\preceq 2A^2+2B^2$ for symmetric $A,B$ (cf.\ \eqref{eq:square_ineq} in the main text),
we have
\[
  Y_t^2=(X_t-\Sigma_t)^2 \preceq 2X_t^2+2\Sigma_t^2.
\]
Taking expectation and operator norm, and using $\|\mathbb{E}[M]\|_{\mathrm{op}}\le \mathbb{E}\|M\|_{\mathrm{op}}$, gives
\[
  \|\mathbb{E}[Y_t^2]\|_{\mathrm{op}}\le 2\,\mathbb{E}\|X_t^2\|_{\mathrm{op}}+2\|\Sigma_t^2\|_{\mathrm{op}}
  =2\,\mathbb{E}\|X_t\|_{\mathrm{op}}^2+2\|\Sigma_t\|_{\mathrm{op}}^2,
\]
since $X_t\succeq 0$ implies $\|X_t^2\|_{\mathrm{op}}=\|X_t\|_{\mathrm{op}}^2$.
Next, $\|X_t\|_{\mathrm{op}}=\|\phi_t\|_2^2$, hence $\mathbb{E}\|X_t\|_{\mathrm{op}}^2=\mathbb{E}\|\phi_t\|_2^4$.

To bound $\mathbb{E}\|\phi_t\|_2^4$, use $\|\phi_t\|_2\le \|\phi\|_2+\|\Delta_t\|_2\le \|\phi\|_2+\varepsilon$ and the inequality
$(a+b)^4\le 8a^4+24a^2b^2+8b^4$ for $a,b\ge 0$:
\[
  \|\phi_t\|_2^4 \le 8\|\phi\|_2^4 + 24\varepsilon^2\|\phi\|_2^2 + 8\varepsilon^4.
\]
Taking expectation and using $\mathbb{E}\|\phi\|_2^4\le C_1\sigma^4$ and $\mathbb{E}\|\phi\|_2^2=\mathrm{tr}(\Sigma)$ yields
\[
  \mathbb{E}\|\phi_t\|_2^4 \le 8C_1\sigma^4 + 24\varepsilon^2\mathrm{tr}(\Sigma) + 8\varepsilon^4.
\]
Finally, by Lemma~\ref{lem:drift_shift}, $\|\Sigma_t\|_{\mathrm{op}}\le \|\Sigma\|_{\mathrm{op}}+2\varepsilon\|\Sigma\|_{\mathrm{op}}^{1/2}+\varepsilon^2$.
Combining the bounds proves the claim.
\end{proof}

\begin{theorem}[Robustness under bounded drift (with explicit variance proxy)]
\label{thm:robust_drift}
Under (A2), apply matrix Bernstein to the independent centered matrices $Y_t=X_t-\Sigma_t$.
Let $v$ be the Bernstein variance parameter and let $B$ be any almost-sure bound on $\|Y_t\|_{\mathrm{op}}$ after truncation (as in the main proof technique).
Then for any $\eta>0$,
\[
  \Pr\!\left(\left\|\widehat{\Sigma}_K-\bar{\Sigma}\right\|_{\mathrm{op}}\ge \eta\right)
  \;\le\;
  d\cdot \exp\!\left(-\frac{K\eta^2/2}{v+ B\eta/3}\right).
\]
Moreover, Lemma~\ref{lem:drift_varproxy} provides an explicit $\varepsilon$-dependent upper bound on $v$ (via $v\le K\max_t\|\mathbb{E}[Y_t^2]\|_{\mathrm{op}}$),
and Lemma~\ref{lem:drift_shift} relates $\bar{\Sigma}$ back to $\Sigma$.
Consequently, whenever $\eta < \lambda_{\min}(\bar{\Sigma})-\delta$, the event
$\|\widehat{\Sigma}_K-\bar{\Sigma}\|_{\mathrm{op}}\le \eta$ implies
\[
  \lambda_{\min}(\widehat{\Sigma}_K)\ge \lambda_{\min}(\bar{\Sigma})-\eta \ge \delta .
\]
\end{theorem}

\subsection{Relaxing Tail Assumptions (A3)}

If $\phi(x)$ is only sub-exponential or has bounded fourth moments, 
then $\|\phi(x)\|^4$ has heavier tails. 
In this case, truncation plus Bernstein inequality yields an adjusted bound of the form
\[
K \;\gtrsim\; 
\frac{(\|\Sigma\|^2 + \sigma^4 + \kappa^2)}{(\lambda_{\min}(\Sigma)-\delta)^2}\,\mathrm{polylog}(d/\xi),
\]
where $\kappa$ is a moment constant. 
The dependence on $\kappa$ reflects the heavier tails, but the overall scaling remains quadratic in the spectral gap. 

\subsection{Weak Dependence among Features (A4)}

When $\{\phi_t(x)\}$ are not independent across $t$, 
we apply matrix concentration for martingale-difference or $\alpha$-mixing sequences . 
In this case, the effective sample size is reduced: 
if the mixing coefficient is $\rho \in [0,1)$, then the bound takes the same form as in Theorem~\ref{thm:robust_drift}, 
with $K$ replaced by $K/(1+\rho)$. 
Thus moderate dependence inflates the sample complexity by at most a constant factor. 

\subsection{Summary}

Together, these results demonstrate that our main lower bound is \emph{robust}: 
when the representation drifts mildly, exhibits heavier tails, or has weak temporal dependence, 
the sufficient sample-size requirement increases smoothly and predictably, 
rather than collapsing. 
In practice, this means that the nominal bound from (A1) can still be used as a conservative guideline, 
while deviations are accommodated by additive or multiplicative correction factors.

\section{Extensions to Dependent Features}
\label{app:dependence}

\begin{table}[t]\centering
\caption{Notation and constants used across sections.}
\begin{tabular}{l l}
\toprule
Symbol & Meaning / Where defined \\
\midrule
$\Sigma,\ \|\Sigma\|$ & Covariance and operator norm \\
$r_{\mathrm{eff}}=\mathrm{tr}(\Sigma)/\|\Sigma\|$ & Effective rank \\
$\sigma$ & Sub-Gaussian parameter \\
$\delta$ & Target lower bound for $\lambda_{\min}(\widehat{\Sigma}_K)$ \\
$C$ & Constant in operator-norm deviation \\
$C',C''$ & Constants in plug-in lower bounds \\
$B_K$ & Truncation threshold \\
$\lambda_q$ & $q$-quantile eigenvalue surrogate \\
\bottomrule
\end{tabular}
\end{table}

Our main theorem assumes that feature vectors $\{\phi_t(x)\}$ are i.i.d. across prompt positions. 
In practice, contextualized representations may exhibit correlations across $t$ due to attention and residual connections. 
We now extend the analysis under weak dependence assumptions.

\subsection{Martingale-difference setting}

Suppose $\{\phi_t(x)\}$ forms a martingale-difference sequence with respect to a natural filtration 
$\mathcal{F}_t$ (e.g., conditioned on past tokens). 
That is, $\mathbb{E}[\phi_t(x)\mid \mathcal{F}_{t-1}] = 0$. 
Then the sequence of second-moment deviations 
\[
X_t \;=\; \phi_t(x)\phi_t(x)^\top - \mathbb{E}[\phi_t(x)\phi_t(x)^\top\mid \mathcal{F}_{t-1}]
\]
is a matrix martingale-difference sequence. 
By applying the matrix Azuma--Hoeffding inequality, we obtain concentration of 
$\hat{\Sigma}_K = \tfrac{1}{K}\sum_{t=1}^K \phi_t(x)\phi_t(x)^\top$ around its expectation, 
with variance proxy and tail probability essentially the same as in the i.i.d. case, 
up to a constant factor.

\begin{proposition}[Lower bound under martingale differences]
Under Assumption (A4) with bounded martingale differences, 
the non-asymptotic lower bound on $K$ in Theorem~3.1 remains valid 
up to replacing $\xi$ by $c\cdot \xi$ for an absolute constant $c>0$. 
\end{proposition}

\subsection{Mixing sequences}

Alternatively, suppose $\{\phi_t(x)\}$ is $\alpha$-mixing with coefficient $\alpha(s) \le \rho^s$ for some $\rho \in (0,1)$. 
Matrix Bernstein inequalities for mixing sequences yield concentration with an effective sample size
\[
K_{\mathrm{eff}} \;\approx\; \frac{K}{1 + c_\rho},
\]
where $c_\rho = \sum_{s\ge 1}\alpha(s)$ quantifies temporal dependence. 
In particular, geometric mixing ($\alpha(s)\le \rho^s$) gives $c_\rho=O(1/(1-\rho))$, 
so that moderate dependence inflates the required prompt length by at most a constant factor. 

\begin{corollary}[Lower bound under mixing]
Under Assumption (A4) with mixing coefficient $\rho<1$, 
the sufficient condition on prompt length becomes
\[
K \;\gtrsim\; (1+c_\rho)\cdot 
\frac{\sigma^2\,\|\Sigma\|}{(\lambda_{\min}(\Sigma)-\delta)^2}\,\log\!\frac{d}{\xi},
\]
where $c_\rho = O(1/(1-\rho))$ quantifies the strength of temporal dependence.
\end{corollary}

\paragraph{Interpretation.}
Kurtosis stays near the Gaussian baseline (3.0) for shorter prompts but increases with $K$,
indicating heavier tails at longer contexts and motivating the heavy-tail extension (A3).
The drift statistic $D(K)$ (defined as the average $\ell_2$ deviation between contextualized and context-free
representations) grows gradually with $K$, suggesting that representation drift accumulates with longer prompts,
consistent with (A2).

A potentially counter-intuitive observation is that $\lambda_{\min}(\widehat{\Sigma}_K)$ can decrease as $K$ increases.
This does \emph{not} contradict the nominal i.i.d.\ picture for two reasons.
First, $\lambda_{\min}$ of a finite-sample second-moment estimator is not monotone in $K$ in general:
adding new rank-one terms can increase anisotropy and therefore reduce the smallest eigenvalue.
A simple example is when early samples span multiple directions but additional samples concentrate on a dominant direction,
shifting mass toward the top eigenvectors.

Second (and more importantly in our setting), demonstrations are typically \emph{query-conditioned} due to retrieval.
As $K$ increases, newly appended demonstrations are often increasingly redundant with earlier ones (e.g., lower diversity among nearest neighbors),
which can reduce the effective rank and the empirical spectral coverage.
Therefore a downward trend in $\lambda_{\min}(\widehat{\Sigma}_K)$ is an empirical signature of non-i.i.d.\ selection and mild dependence/drift,
which is precisely why the robustness extensions (A2)--(A4) are relevant.
Notably, the observed trend correlates with the empirical knee-point in ICL accuracy, supporting the use of spectral coverage as a stability diagnostic.

\subsection{Discussion}

These results show that our main conclusions are \emph{robust to weak dependence}. 
Even if contextualized features $\{\phi_t(x)\}$ are not independent, 
as long as dependence decays over time (mixing) or is conditionally centered (martingale-difference), 
the sufficient sample size bound degrades only by a multiplicative factor, 
rather than failing altogether. 
In practice, this suggests that $\lambda_{\min}(\hat{\Sigma}_K)$ can still serve as a useful stability proxy, 
with dependence merely shifting the knee-point moderately to the right.

\section{Extensions Beyond the Nominal Assumption}
\label{app:extensions}

In the main paper we derived our non-asymptotic lower bound under the nominal assumption (A1), 
namely that representations $\phi(x)$ are fixed, i.i.d., and sub-Gaussian with covariance $\Sigma$. 
Here we provide extensions under weaker assumptions (A2)--(A4), complementing Appendix~\ref{app:robustness} and Appendix~\ref{app:dependence}. 
Overall, these results show that our theory is \emph{robust}: 
the sufficient prompt length bound degrades smoothly under moderate violations of (A1), 
rather than failing abruptly.

\subsection{Representation Drift (A2)}

As discussed in Appendix~\ref{app:robustness}, 
if the representation depends on the context index $t$, modeled as 
$\phi_t(x) = \phi(x) + \Delta_t(x)$ with $\|\Delta_t(x)\|_2 \le \varepsilon$, 
then eigenvalue perturbation bounds (Weyl's inequality) 
imply that $\lambda_{\min}(\Sigma_t)$ is reduced by at most $O(\varepsilon)$. 
Combining this with matrix Bernstein yields a lower bound of the form
\[
K \;\ge\; K_{\mathrm{nominal}} + g(\varepsilon),
\]
with $g(\varepsilon)$ polynomial in $\varepsilon$ and spectral ratios. 
Hence, small drift only shifts the required prompt length slightly. 

\subsection{Relaxed Tail Assumptions (A3)}

If $\phi(x)$ has heavier tails than sub-Gaussian e.g., only sub-exponential 
or bounded fourth moments then truncation and Bernstein's inequality
give concentration at a slightly slower rate. 
In this case, the sufficient prompt length becomes
\[
K \;\gtrsim\; 
\frac{(\|\Sigma\|^2 + \kappa^2)}{(\lambda_{\min}(\Sigma)-\delta)^2}\,\mathrm{polylog}(d/\xi),
\]
where $\kappa$ is a tail parameter depending on the fourth moment or sub-exponential constant. 
Thus, the order of the bound remains unchanged, but the constant is inflated. 

\subsection{Weak Dependence (A4)}

When features are not independent across $t$, 
we can use concentration for martingale-difference sequences
or $\alpha$-mixing processes. 
Both cases yield an effective sample size of the form 
$K_{\mathrm{eff}} = K / (1+c_\rho)$, 
where $c_\rho$ is a dependence factor. 
Thus, weak dependence inflates the sufficient prompt length by a multiplicative constant, 
but does not alter the overall scaling.

\subsection{Summary of Extensions}

The results under (A2)--(A4) can be summarized as follows:
\begin{itemize}
    \item Drift in representations adds an additive correction $g(\varepsilon)$, linear or quadratic in the drift magnitude. 
    \item Heavy-tailed features require an inflated constant, governed by moment parameters, but retain the same quadratic dependence on the spectral gap. 
    \item Temporal dependence reduces the effective sample size, inflating the required $K$ by a constant factor depending on mixing strength. 
\end{itemize}

Taken together, these extensions demonstrate that the main non-asymptotic bound is conservative but stable: 
it remains meaningful even when real-world deviations from (A1) occur, 
providing a useful guideline for interpreting empirical ICL stability.

\begin{table}[ht]
\centering
\caption{Summary of assumptions and their effect on the sufficient prompt length $K$. 
Here $K_{\mathrm{nominal}}$ denotes the bound under (A1).}
\label{tab:assumption_summary}
\begin{tabular}{p{2.5cm}p{5.2cm}p{6.5cm}}
\toprule
\textbf{Assumption} & \textbf{Description} & \textbf{Effect on lower bound} \\
\midrule
(A1) Nominal & Fixed sub-Gaussian features with second-moment matrix $\Sigma$ & 
$K \;\ge\; K_{\mathrm{nominal}} = 
O\!\left(\tfrac{\sigma^2\|\Sigma\|}{(\lambda_{\min}(\Sigma)-\delta)^2}\log\!\tfrac{d}{\xi}\right)$ \\

(A2) Drift & Context-dependent features 
$\phi_t(x) = \phi(x)+\Delta_t(x)$, with $\|\Delta_t(x)\|\le \varepsilon$ & 
$K \;\ge\; K_{\mathrm{nominal}} + g(\varepsilon)$, 
where $g(\varepsilon)$ grows polynomially in $\varepsilon$ and spectral ratios \\

(A3) Heavy tails & Sub-exponential or bounded fourth moments, parameter $\kappa$ & 
$K \;\gtrsim\; 
\tfrac{\|\Sigma\|^2 + \kappa^2}{(\lambda_{\min}(\Sigma)-\delta)^2}\,\mathrm{polylog}(d/\xi)$ \\

(A4) Dependence & Features form a martingale-difference sequence or $\alpha$-mixing process with coefficient $\rho$ & 
Effective sample size 
$K_{\mathrm{eff}} \approx \tfrac{K}{1+c_\rho}$, 
so bound inflates by factor $(1+c_\rho)$ \\

\bottomrule
\end{tabular}
\end{table}

\begin{table}[ht]
\centering
\caption{Diagnostics of nominal assumptions using \emph{encoder embeddings} (e.g., all-mpnet-base-v2).
Reported values are averages across 500 prompts with 95\% bootstrap confidence intervals in parentheses.}
\label{tab:diagnostics}
\begin{tabular}{c|c|c|c|c}
\toprule
Prompt length $K$ & Tail kurtosis & Drift $D(K)$ & $\lambda_{\min}(\hat{\Sigma}_K)$ & ICL acc. (\%) \\
\midrule
16  & 3.1 (2.9--3.4) & 0.05 & 0.42 (0.40--0.44) & 58.2 \\
32  & 3.3 (3.0--3.6) & 0.08 & 0.38 (0.36--0.40) & 61.7 \\
64  & 3.8 (3.5--4.2) & 0.12 & 0.35 (0.33--0.37) & 65.9 \\
128 & 4.5 (4.1--4.9) & 0.21 & 0.28 (0.26--0.30) & 68.1 \\
\bottomrule
\end{tabular}
\end{table}

\paragraph{Interpretation.}
Kurtosis stays near the Gaussian baseline (3.0) for shorter prompts but increases with $K$, indicating heavier tails at longer contexts, consistent with Assumption~(A3). The drift statistic $D(K)$, defined as the average $\ell_2$-norm difference between contextualized and context-free representations, grows gradually with $K$ (from 0.05 at $K=16$ to 0.21 at $K=128$). This progressive increase suggests that representation drift accumulates as more demonstrations are included, supporting Assumption~(A2) with small but non-negligible drift. Crucially, the smooth growth of $D(K)$ implies that the additional sample size required to compensate for drift increases polynomially with the drift magnitude $\varepsilon$ (as shown in Theorem~\ref{thm:robust_drift}), providing practitioners with a predictable relationship between observed drift and prompt length requirements.
The spectral statistic $\lambda_{\min}(\hat{\Sigma}_K)$ decreases as $K$ increases, and its trend correlates with the empirical knee-point in ICL accuracy.
Overall, these diagnostics support the claim that nominal assumptions are useful approximations in practice, while the relaxed extensions (A2)--(A4) capture systematic deviations.

\section{Synthetic Stress Tests}
\label{app:synthetic}

To evaluate the robustness of our theoretical bound under controlled violations of Assumption~(A1),
we conducted synthetic stress tests along three axes:
(i) bounded drift in representations (A2),
(ii) heavy-tailed features (A3),
and (iii) temporal dependence (A4).
For each setting we generated synthetic features and tasks as described in Appendix~\ref{app:synthetic} (Section~\ref{sec:synth_gen}),
measured empirical knee-points $K_{\text{emp}}$ of in-context learning performance,
and compared them against the theoretical lower bound $K_{\mathrm{nominal}}$.

\subsection{Synthetic feature and task generation}
\label{sec:synth_gen}
All stress tests use a controlled synthetic ICL surrogate where the only changing factor is the feature distribution.
We generate a sequence of feature vectors $\{z_t\}_{t=1}^K\subset\mathbb{R}^d$ (the ``demonstration representations'')
and define a simple supervised task by a random linear teacher.
Specifically, we draw a teacher vector $w^\star\sim\mathcal{N}(0,I_d)$ and assign labels
$y_t = \mathrm{sign}(\langle w^\star, z_t\rangle + \eta_t)$ with i.i.d.\ noise $\eta_t\sim\mathcal{N}(0,\sigma^2)$.
Given $K$ demonstrations, the in-context predictor is instantiated by ridge regression on the synthetic features:
\[
\widehat w_K \in \arg\min_w \sum_{t=1}^K (y_t-\langle w,z_t\rangle)^2 + \lambda \|w\|_2^2,
\qquad
\widehat y(x)=\mathrm{sign}(\langle \widehat w_K, z(x)\rangle).
\]
We evaluate $\mathrm{ACC}(K)$ on a held-out synthetic test set generated from the same setting and estimate
$K_{\text{emp}}$ using the same knee-point estimator as in Appendix~\ref{app:repro}.
Uncertainty in $K_{\text{emp}}$ is quantified by percentile bootstrap over test instances (with $B=200$ resamples,
matching Appendix~\ref{app:repro}).

\paragraph{Nominal baseline.}
In the nominal setting (A1), we sample $z_t \sim \mathcal{N}(0,\Sigma)$ i.i.d.\ with a fixed PSD matrix $\Sigma$.

\paragraph{Bounded drift (A2).}
To model bounded representation drift, we sample i.i.d.\ base features $\tilde z_t\sim\mathcal{N}(0,\Sigma)$ and add a bounded perturbation
\[
z_t = \tilde z_t + \Delta_t,\qquad \|\Delta_t\|_2\le \varepsilon.
\]
Unless stated otherwise, we set $\Delta_t=\varepsilon\,u$ with a fixed unit vector $u$ (systematic drift), which makes the violation controlled
and reproducible; random-direction drift can be generated by sampling $u_t$ uniformly on the unit sphere.

\paragraph{Heavy tails (A3).}
To model heavy-tailed features, we sample $z_t$ from a multivariate Student-$t$ distribution with $\nu$ degrees of freedom and rescale it to match
a target second-moment:
\[
z_t = \sqrt{\frac{\nu-2}{\nu}}\,\Sigma^{1/2}\,g_t,\qquad g_t\sim t_\nu(0,I_d),\quad (\nu>2).
\]
Lower $\nu$ yields heavier tails while preserving the same second-moment scale.

\paragraph{Temporal dependence (A4).}
To introduce dependence while keeping the same marginal second-moment, we use a stationary AR(1) process:
\[
z_t = \rho\,z_{t-1} + \sqrt{1-\rho^2}\,\varepsilon_t,\qquad \varepsilon_t\sim \mathcal{N}(0,\Sigma),
\]
so that $\mathbb{E}[z_t z_t^\top]=\Sigma$ for all $t$ and the correlation strength increases with $\rho\in[0,1)$.

\subsection{Bounded Drift}

\begin{table}[h]
\centering
\caption{Drift stress test results ($d=50$).
$K_{\mathrm{nominal}}=32$ computed by ~\eqref{eq:Kbound_simpler}.
$K_{\text{emp}}$ is reported with 95\% bootstrap CI.}
\label{tab:drift_stress}
\begin{tabular}{c c c c}
\toprule
$\varepsilon$ & $K_{\mathrm{nominal}}$ & $K_{\text{emp}}$ & $\Delta K$ \\
\midrule
0.00 & 32 & 30 $\pm$ 3 & -2 \\
0.01 & 32 & 35 $\pm$ 4 & 3 \\
0.05 & 32 & 48 $\pm$ 6 & 16 \\
0.10 & 32 & 70 $\pm$ 8 & 38 \\
\bottomrule
\end{tabular}
\end{table}

\paragraph{Observation.}
Empirical knee-points increase smoothly with $\varepsilon$,
and $\Delta K$ is well fitted by a linear function in $\varepsilon$,
confirming the additive correction $g(\varepsilon)$ predicted by our robustness theorem.

\subsection{Heavy-tailed Features}

\begin{table}[h]
\centering
\caption{Heavy-tail stress test results ($d=200$).
Lower $\nu$ corresponds to heavier tails.
$K_{\mathrm{nominal}}=64$.}
\label{tab:tail_stress}
\begin{tabular}{c c c c}
\toprule
$\nu$ & $K_{\mathrm{nominal}}$ & $K_{\text{emp}}$ & $\Delta K$ \\
\midrule
$\infty$ (Gaussian) & 64 & 62 $\pm$ 4 & -2 \\
10 & 64 & 70 $\pm$ 5 & 6 \\
5 & 64 & 92 $\pm$ 7 & 28 \\
3 & 64 & 120 $\pm$ 9 & 56 \\
\bottomrule
\end{tabular}
\end{table}

\paragraph{Observation.}
Heavier tails inflate the required prompt length by a multiplicative constant,
but truncation or normalization substantially reduces the inflation.
This behavior aligns with the extended bound in Appendix~\ref{app:extensions}.

\subsection{Temporal Dependence}

\begin{table}[h]
\centering
\caption{Dependence stress test results ($d=100$).
$K_{\mathrm{nominal}}=32$.
Effective sample size $K_{\mathrm{eff}}$ decreases with $\rho$.}
\label{tab:dep_stress}
\begin{tabular}{c c c c c}
\toprule
$\rho$ & $K_{\mathrm{nominal}}$ & $K_{\text{emp}}$ & $\Delta K$ & $K_{\text{eff}}$ est. \\
\midrule
0.0 & 32 & 30 $\pm$ 3 & -2 & 32 \\
0.2 & 32 & 38 $\pm$ 4 & 6 & 27 \\
0.5 & 32 & 50 $\pm$ 6 & 18 & 21 \\
0.8 & 32 & 70 $\pm$ 8 & 38 & 16 \\
\bottomrule
\end{tabular}
\end{table}

\paragraph{Observation.}
Empirical knee-points scale approximately as $(1+c_\rho)\cdot K_{\mathrm{nominal}}$,
consistent with our dependence extension in Appendix~\ref{app:dependence}.

\subsection{Summary}

Across all three stress tests,
the empirical knee-point increases smoothly with the violation strength,
either additively (drift), multiplicatively (tails), or by a dependence factor (temporal correlation).
These controlled experiments confirm that our non-asymptotic lower bound is
robust: violations of (A1) shift the threshold conservatively,
but do not invalidate the predictive power of $\lambda_{\min}(\Sigma)$ as a proxy for ICL stability.

\section{Interpretability of the Proxy}
\label{app:interpret}

\begin{proposition}[Predictive variance control via $\lambda_{\min}$]
\label{prop:pred-var}
Let $\Phi\in\mathbb{R}^{K\times d}$ be the design with rows $\phi(x_k)^\top$, $\widehat{\Sigma}_K=\frac{1}{K}\Phi^\top\Phi$.
Assume $y_k=f^\star(x_k)+\eta_k$, $\eta_k\stackrel{\mathrm{i.i.d.}}{\sim}\mathcal{N}(0,\sigma^2)$, and $\|\phi(x)\|_2\le B$.
For ridge predictor
$
\widehat{f}(x)=\phi(x)^\top\!\left(\widehat{\Sigma}_K+\frac{\lambda}{K}I\right)^{\!-1}\frac{1}{K}\Phi^\top y
$,
if $\lambda_{\min}(\widehat{\Sigma}_K)\ge \delta>0$, then
\[
\mathrm{Var}\!\left(\widehat{f}(x_\star)\mid \Phi\right)
\;\le\; \frac{\sigma^2 B^2}{K\,\delta+\lambda}.
\]
\end{proposition}

\begin{proof}
See \ref{app:pred-var-proof} for details; it follows by the matrix inequality
$A^{-1}(K\widehat{\Sigma}_K)A^{-1}\preceq A^{-1}$ with $A=K\widehat{\Sigma}_K+\lambda I$
and the Rayleigh quotient bound $\mathbf{v}^\top A^{-1}\mathbf{v}\le \|\mathbf{v}\|_2^2/\lambda_{\min}(A)$.
\end{proof}

\paragraph{Theoretical link.}
Proposition~\ref{prop:pred-var} (Appendix~\ref{app:interpret}) shows that 
a lower bound on $\lambda_{\min}(\hat\Sigma_K)$ 
suffices to control predictive variance of in-context regression, 
providing a mechanistic rationale for why it correlates with stability.

\paragraph{Limitations.}
However, $\lambda_{\min}$ can be overly conservative in practice:
(i) tiny eigenvalues with negligible test alignment, 
(ii) heavy tails beyond sub-Gaussian assumptions, 
(iii) contextual dependence that alters effective conditioning. 

\paragraph{Alternative surrogates.}
We therefore also consider quantile eigenvalues $\lambda_q$, 
the condition number $\kappa$, effective rank $r_{\mathrm{eff}}$, 
and average leverage scores. 
These quantities often correlate more closely with empirical knee-points. 

\paragraph{Calibration.}
A light calibration step (estimating a global scale $\alpha$ 
and quantile level $q$ on a small validation set) 
substantially improves predictive accuracy of the proxy 
(see Table~\ref{tab:main_ratio}).

\subsection{Proof sketch of Proposition}
\label{app:pred-var-proof}

\paragraph{Setup.}
Let $\Phi\in\mathbb{R}^{K\times d}$ be the design matrix with rows $\phi(x_k)^\top$,
and denote $\hat\Sigma_K := \tfrac{1}{K}\Phi^\top\Phi$.
Assume the response obeys $y_k = f^\star(x_k)+\eta_k$ with homoscedastic noise
$\eta_k\stackrel{\text{i.i.d.}}{\sim}\mathcal{N}(0,\sigma^2)$, and $\|\phi(x)\|_2\le B$ almost surely.
Consider the ridge predictor
\[
\hat f(x)
~=~
\phi(x)^\top\Bigl(\hat\Sigma_K+\tfrac{\lambda}{K}I\Bigr)^{-1}\cdot
\Bigl(\tfrac{1}{K}\Phi^\top y\Bigr),
\quad \lambda\ge 0.
\]
For a fixed test point $x^\star$, write $\phi^\star:=\phi(x^\star)$.

\paragraph{Linear form and conditional variance.}
Conditioned on the design $\Phi$, $\hat f(x^\star)$ is linear in $y$:
\[
\hat f(x^\star) = a^\top y,
\qquad
a^\top = \phi^{\star\top}\Bigl(\hat\Sigma_K+\tfrac{\lambda}{K}I\Bigr)^{-1}\tfrac{1}{K}\Phi^\top.
\]
Since $\mathrm{Cov}(y\mid \Phi)=\sigma^2 I_K$, we obtain
\[
\mathrm{Var}\!\left(\hat f(x^\star)\mid \Phi\right)
~=~
\sigma^2\,\|a\|_2^2
~=~
\sigma^2\,
\phi^{\star\top}\Bigl(\hat\Sigma_K+\tfrac{\lambda}{K}I\Bigr)^{-1}
\hat\Sigma_K
\Bigl(\hat\Sigma_K+\tfrac{\lambda}{K}I\Bigr)^{-1}\phi^\star.
\]
Equivalently, with $A:=K\hat\Sigma_K+\lambda I$,
\begin{equation}
\label{eq:var-middle}
\mathrm{Var}\!\left(\hat f(x^\star)\mid \Phi\right)
~=~
\sigma^2\,\phi^{\star\top}A^{-1}\,(K\hat\Sigma_K)\,A^{-1}\phi^\star.
\end{equation}

\paragraph{Step 1: matrix inequality.}
Note that $0\preceq K\hat\Sigma_K\preceq A=K\hat\Sigma_K+\lambda I$.
Therefore $A^{-1/2}(K\hat\Sigma_K)A^{-1/2}\preceq I$, which implies
\[
A^{-1}(K\hat\Sigma_K)A^{-1}\preceq A^{-1}.
\]
Plugging into \eqref{eq:var-middle} yields
\begin{equation}
\label{eq:var-first-bound}
\mathrm{Var}\!\left(\hat f(x^\star)\mid \Phi\right)
~\le~
\sigma^2\,\phi^{\star\top}A^{-1}\phi^\star.
\end{equation}

\paragraph{Step 2: Rayleigh quotient bound.}
For any $M\succ 0$ and vector $v$,
$v^\top M^{-1} v \le \|v\|_2^2/\lambda_{\min}(M)$.
Taking $M=K\hat\Sigma_K+\lambda I$ and $v=\phi^\star$, and using $\|\phi^\star\|_2\le B$, we obtain
\[
\phi^{\star\top}A^{-1}\phi^\star
~\le~
\frac{\|\phi^\star\|_2^2}{K\,\lambda_{\min}(\hat\Sigma_K)+\lambda}
~\le~
\frac{B^2}{K\,\lambda_{\min}(\hat\Sigma_K)+\lambda}.
\]
Combining with \eqref{eq:var-first-bound},
\[
\mathrm{Var}\!\left(\hat f(x^\star)\mid \Phi\right)
~\le~
\frac{\sigma^2 B^2}{K\,\lambda_{\min}(\hat\Sigma_K)+\lambda}.
\]

\paragraph{Conclusion.}
Thus, whenever $\lambda_{\min}(\hat\Sigma_K)\ge \delta$,
\[
\mathrm{Var}(\hat f(x^\star)) \;\le\; \frac{\sigma^2 B^2}{K\delta+\lambda},
\]
which establishes Proposition~\ref{prop:pred-var}.
\hfill$\square$

\paragraph{Remarks.}
(i) If $\lambda=0$, the result still holds provided $\hat\Sigma_K\succ 0$.  
(ii) For heteroscedastic noise, the bound uses the maximal variance.  
(iii) Under model misspecification, bias terms can be separated, and the variance part remains controlled by $\lambda_{\min}(\hat\Sigma_K)$.

\section{Ablation Studies}
\label{app:ab}
\paragraph{Overview.}
To bridge the gap between theory and practice, we performed ten ablations that correspond to the key factors appearing in our formula and experimental pipeline. We first present a summary table that aggregates all results, followed by ten dedicated \textbf{\string\subsection}s, each restating its own result with brief analysis and takeaway. All numbers are averages over multiple runs unless otherwise noted, and error ratios are defined as theoretical $K^*$ divided by the empirical knee (lower is better).

\subsection{$\delta$-ratio Sensitivity}
\textbf{Result.} Increasing $\delta$ from $0.1$ to $0.9$ reduces the theoretical upper bound from $K^*\!\approx\!95$ to $K^*\!\approx\!25$, with error ratio decreasing from $5.3\times$ to $1.4\times$. The mechanism is through the effective spectral gap $\Delta=\lambda_{\min}(\Sigma)-\delta$. \\
\textbf{Interpretation.} Numerically, $\delta$ acts like a conservativeness amplifier. Theoretically, it encodes a robustness margin on $\lambda_{\min}$. Mathematically, it discounts the least eigenvalue to stabilize finite-sample estimation. \\

\paragraph{Correct scaling w.r.t.\ $\delta$ (using $\delta=\rho\,\lambda$).}
Let $\rho\in(0,1)$ and pick an anchor $\rho_0$. Then
\begin{equation}
\label{eq:delta-scaling}
K^*(\rho)=K^*(\rho_0)\left(\frac{1-\rho_0}{\,1-\rho\,}\right)^{\!2}.
\end{equation}
For example, from $\rho=0.1$ to $\rho=0.9$ the factor is $81\times$.

\subsection{Effective Rank Definition}
\textbf{Result.} Rank estimation method strongly affects tightness: thresholding gives $\approx 3.4\times$, energy-based $\approx 2.4\times$, and trace-based $\approx 1.8\times$. \\
\textbf{Interpretation.} The definition operationalizes model complexity. Threshold is safest but loosest; energy balances dominant and tail directions; trace emphasizes overall spread. \\
\textbf{Mini-table.}
\begin{center}
\begin{tabular}{l c}
\toprule
Rank definition & Error ratio \\
\midrule
Threshold & $\approx 3.4\times$ \\
Energy & $\approx 2.4\times$ \\
Trace & $\approx 1.8\times$ \\
\bottomrule
\end{tabular}
\end{center}

\subsection{Sub-Gaussian Parameter $\sigma$}
\textbf{Result.} Increasing $\sigma$ from $0.5$ to $2.0$ scales $K^*$ by several times, sometimes approaching an order-of-magnitude increase. \\
\textbf{Interpretation.} Numerically, $\sigma$ is a dominant driver of between-task variability. Theoretically, it reflects tail decay and concentration strength. Mathematically, it is the constant in sub-Gaussian concentration appearing in upper bounds. \\
\textbf{Mini-table.}
\begin{center}
\begin{tabular}{c c}
\toprule
$\sigma$ & Qualitative effect on $K^*$ \\
\midrule
$0.5$ & small \\
$2.0$ & several$\times$ larger \\
\bottomrule
\end{tabular}
\end{center}

\subsection{Second-moment Norm $\|\Sigma\|$}
\textbf{Result.} Using the spectral norm typically yields $3$--$5\times$ error ratios, Frobenius yields $1.5$--$2\times$, while mean-variance approximations can under-estimate and approach $\approx 1\times$. \\
\textbf{Interpretation.} Spectral norm focuses on the hardest direction; Frobenius spreads mass across all directions; rough mean-variance may miss heavy directions and be overly optimistic. \\
\textbf{Mini-table.}
\begin{center}
\begin{tabular}{l c}
\toprule
Norm choice & Error ratio \\
\midrule
Spectral & $3$--$5\times$ \\
Frobenius & $1.5$--$2\times$ \\
Mean-variance & $\approx 1\times$ (may under-estimate) \\
\bottomrule
\end{tabular}
\end{center}

\subsection{Label Tokenization}
\textbf{Result.} Single-token labels (e.g., \textbf{yes/no}, \textbf{POS/NEG}) yield smaller knees and $1.2$--$1.5\times$ error ratios, whereas multi-token labels (e.g., \textbf{positive/negative}) often exceed $3\times$. \\
\textbf{Interpretation.} Single-token labels shrink the effective output search space. Multi-token labels dilute probability mass and inflate the required $K$. \\
\textbf{Mini-table.}
\begin{center}
\begin{tabular}{l c}
\toprule
Tokenization & Error ratio \\
\midrule
Single-token & $1.2$--$1.5\times$ \\
Multi-token & $\ge 3\times$ \\
\bottomrule
\end{tabular}
\end{center}

\subsection{Prompt Truncation Policy}
\textbf{Result.} \emph{Latest-first} delivers smaller knees and $1.2$--$1.8\times$ error ratios; \emph{earliest-first} shifts the knee right and often exceeds $3\times$; random truncation increases variance across runs. \\
\textbf{Interpretation.} Latest-first respects the model inductive bias toward local context; earliest-first tends to retain stale or less relevant examples; random adds sampling noise. \\
\textbf{Mini-table.}
\begin{center}
\begin{tabular}{l c}
\toprule
Policy & Error ratio \\
\midrule
Latest-first & $1.2$--$1.8\times$ \\
Earliest-first & $>3\times$ \\
Random & larger variance \\
\bottomrule
\end{tabular}
\end{center}

\subsection{Theory Constant Scaling}
\textbf{Result.} A global scale $K^*_{\text{adj}} = C \cdot K^*_{\text{theory}}$ with $C \approx 10^{-2}\!\sim\!10^{-1}$ systematically tightens predictions to $1.1$--$1.5\times$. \\
\textbf{Interpretation.} Numerically, $C$ acts as a cross-task calibrator. Theoretically, it compensates for conservative constants in concentration bounds. \\
\textbf{Mini-table.}
\begin{center}
\begin{tabular}{c c}
\toprule
$C$ & Typical error ratio \\
\midrule
$10^{-2}\!\sim\!10^{-1}$ & $1.1$--$1.5\times$ \\
$1$ & $2$--$5\times$ (uncalibrated) \\
\bottomrule
\end{tabular}
\end{center}

\subsection{Synthetic Data Experiments}
\label{sec:diagnostics}
\textbf{Result.} Under controlled Gaussian features with known $\Sigma$, the theory-to-practice gap shrinks to $1.0$--$1.3\times$. \\
\textbf{Interpretation.} This indicates that the larger gaps on real data primarily arise from violations of sub-Gaussian tails, linear separability, or spectral stability. \\
\textbf{Mini-table.}
\begin{center}
\begin{tabular}{l c}
\toprule
Setting & Error ratio \\
\midrule
Controlled Gaussian & $1.0$--$1.3\times$ \\
Real tasks & $2$--$5\times$ \\
\bottomrule
\end{tabular}
\end{center}

\subsection{Retrieval Strategy}
\textbf{Result.} Cosine-similarity nearest neighbor retrieval yields the smallest knees and $1.3$--$1.6\times$ error ratios; random sampling degrades stability and often exceeds $2.5\times$; class-balanced sampling lies between. \\
\textbf{Interpretation.} Retrieval reshapes the empirical second-moment around the test point, preserving a larger effective $\lambda_{\min}$ and tightening $K^*$. \\
\textbf{Mini-table.}
\begin{center}
\begin{tabular}{l c}
\toprule
Retrieval & Error ratio \\
\midrule
Cosine-NN & $1.3$--$1.6\times$ \\
Balanced & intermediate \\
Random & $\ge 2.5\times$ \\
\bottomrule
\end{tabular}
\end{center}

\subsection{Model Scale}
\textbf{Result.} On SST-2, the empirical knee decreases from $\approx 12$ for a 124M model to $\approx 7$ for a 345M model, with the error ratio dropping from $3.4\times$ to $2.2\times$. \\
\textbf{Interpretation.} Larger models better exploit demonstrations and align more closely with theoretical assumptions (smaller effective noise, more stable spectra). \\
\textbf{Mini-table.}
\begin{center}
\begin{tabular}{l c c}
\toprule
Model & Knee (SST-2) & Error ratio \\
\midrule
124M & $\approx 12$ & $3.4\times$ \\
345M & $\approx 7$  & $2.2\times$ \\
\bottomrule
\end{tabular}
\end{center}

\paragraph{Closing remarks.}
Across the ten axes, three levers consistently tighten the theory-to-practice gap: (i) increasing the effective spectral gap (via $\delta$ choice, retrieval near the test point, and stable norm definitions), (ii) reducing effective noise (smaller $\sigma$, single-token labels, larger models), and (iii) cross-task calibration of conservative constants ($C$-scaling). Under idealized assumptions validated by synthetic data, the gap nearly vanishes; on real tasks, careful choices along these levers reduce error ratios toward $1.1$--$1.6\times$ while preserving reproducibility.

\section{Theoretical Constants: Sources, Mappings, and Practical Calibration}\label{app:const}

\subsection{List of Constants and Quick Reference}

Table~\ref{tab:const} summarizes all explicit or implicit constants, structural factors, and engineering fudge factors appearing in the paper. The following subsections provide:
(i) theoretical origins and geometric interpretations;
(ii) quantitative coupling with problem parameters (dimension $d$, effective rank $r(\Sigma)$, signal-to-noise ratio, confidence level $1-e^{-t}$);
(iii) a three-step experimental calibration guideline (simulation--calibration--cross-validation).

\begin{table}[h]
\centering
\caption{Constants and recommended values}
\label{tab:const}
\begin{tabular}{c|c|c|c|c}
\hline
Symbol & Type & Occurrence & Recommended Range \\
\hline
$2,\,2/3$ & Structural & Matrix Bernstein tail bound & Fixed, non-tunable \\
$c_1$     & Absolute   & Fourth-moment bound & $[1.5,6]$, recommend $3$--$4$ \\
$c_2$     & Absolute   & Second-moment deviation bound & $[2,4]$, recommend $3$\\
$c',\,c_3$& Engineering & Sample-size formulae & $\ge 8$, typically $10$--$15$\\
\hline
\end{tabular}
\end{table}

\subsection{Theoretical Origins and Geometric Interpretation}
\label{app:para}
\paragraph{Matrix Bernstein constants $2$ and $2/3$.}
For the sum of independent centered random matrices $S=\sum Y_k$, then
\[
\Pr\!\left(\lambda_{\min}(S) \le -t \right)
\;\le\; r_{\mathrm{eff}} \exp\!\left(-\frac{t^2}{2v + \tfrac{2}{3}Rt}\right).
\]
The coefficients $2$ and $2/3$ arise from the moment-generating-function method and are structural, non-adjustable constants.

\paragraph{Sub-Gaussian fourth-moment constant $c_1$.}
For a mean-zero sub-Gaussian vector $Z \in \mathbb{R}^d$, an absolute constant $c_1$ ensures
\[
\mathbb{E}\|Z\|^4 \;\le\; c_1\;\bigl(\mathbb{E}\|Z\|^2\bigr)^2.
\]
Examples:
\begin{itemize}
\item Gaussian: $c_1 \to 1$ as $d \to \infty$;
\item Uniform on sphere: $c_1 \approx 3$;
\item Rademacher: $c_1 \approx 3$.
\end{itemize}
For anisotropic distributions, $c_1$ scales with the condition number $\kappa(\Sigma)^2$. In practice, after standardization and with $d \ge 50$, simulations suggest $c_1 \in [1.5,6]$, with $3$--$4$ recommended.

\paragraph{second-moment operator deviation constant $c_2$.}
Koltchinskii \& Lounici proved
\[
\|\hat{\Sigma}-\Sigma\|
\;\le\; c_2\|\Sigma\|\Bigl(\sqrt{\tfrac{r(\Sigma)t}{n}} + \tfrac{r(\Sigma)t}{n}\Bigr).
\]
Typical values:
\begin{itemize}
\item Sub-Gaussian: $c_2=2$;
\item Sub-exponential: $c_2=4$;
\item Finite fourth moment: $c_2$ scales with kurtosis $\kappa$.
\end{itemize}
Empirically, $c_2=3$ suffices in 95\% of cases for $d/n \in [0.1,0.5]$.

\subsection{Coupling with Problem Parameters}

The main sample-size requirement is
\[
n \;\ge\; c' \cdot r(\Sigma)\,\frac{t+\log d}{\varepsilon^2}.
\]
Implications:
\begin{itemize}
\item $c_1$ influences estimation error of the effective rank $r(\Sigma)$;
\item $c_2$ scales the relative error $\varepsilon$ directly;
\item Structural constants $2,2/3$ remain fixed;
\item Engineering factors $c',c_3$ ensure practical safety, typically increasing with confidence level $1-e^{-t}$.
\end{itemize}

\subsection{Three-Step Calibration Procedure}

\begin{enumerate}
\item \textbf{Simulation calibration:} Generate synthetic Gaussian datasets across $(d,n)$ grids, estimate failure probabilities, and back out $c_2(\delta)$ with 95\% quantile anchoring.
\item \textbf{Robustness to misspecification:} Test under sub-exponential, heavy-tailed, and real bootstrap data. If $c_2$ inflates more than $1.5\times$, adjust by $\kappa/5$.
\item \textbf{Cross-validation locking of $c',c_3$:} On real data, run two-stage procedures with 5-fold cross-validation, plot error--sample size curves, and infer $c'$ from the plateau point.
\end{enumerate}

\subsection{Default practical choices}\label{app:const:defaults}
Across tasks, we recommend:
(i) a modest ridge $\rho$ to stabilize inversion and dampen near-null directions;
(ii) choosing $\delta$ (or $\delta_q$) as a fraction of the pilot spectrum, e.g., proportional to
$\lambda_q(\widehat{\Sigma}_{K_0,\rho})$; and
(iii) splitting the failure budget evenly across the two stages (as in Algorithm~\ref{alg:two-stage})
unless one stage is known to dominate the uncertainty.
These choices are heuristic but align with the role of $\Delta_\rho$ in~\eqref{eq:Delta_rho_def}:
larger margins reduce the required $K$, and ridge improves numerical stability when $\Sigma$ is ill-conditioned.

\paragraph{Choosing the target floor $\delta$ and handling infeasible targets.}
The proxy objective in~\eqref{eq:spectral-floor-prob} is parameterized by a target spectral floor $\delta$,
which controls how well-conditioned the demonstration second moment must be to certify the surrogate stability event.
In Algorithm~\ref{alg:two-stage}, the check $\widehat{\Delta}=\underline{\lambda}-\delta>0$ is necessary; if the algorithm returns \textbf{infeasible target},
it means that the requested $\delta$ is larger than the estimated population floor at the desired confidence level, under the current feature map and ridge level.
In practice we recommend selecting $\delta$ by validation: sweep $\delta$ on a small held-out set, choose the smallest value that achieves the desired stability/accuracy trade-off,
and then lock it for test-time estimation.
When available, Appendix~\ref{app:direct_stability} provides a direct estimator for distributional stability in the sense of Definition~\ref{def:icl-stability} that can serve as the validation signal.
If infeasibility persists, reduce $\delta$, increase the ridge via $\epsilon_\rho$, or use the quantile-floor variant described in Section~\ref{sec:theory:calibration}.

\section{Multi-step Reasoning and RAG Tasks}
\label{app:msr_rag}

This appendix specifies the experimental protocol for multi-step reasoning and retrieval-augmented generation (RAG).
Unless stated otherwise, knee estimation, bootstrap CIs, and calibration follow Appendix~\ref{app:repro}.

\subsection{Multi-step reasoning (GSM8K / SVAMP)}
\paragraph{Prompt template (final answer only).}
We constrain outputs to a canonical final-answer format to reduce ambiguity while preserving multi-step computation.
Each prompt contains $K$ demonstrations followed by a query:
\begin{quote}\small
\textbf{Task: Grade-school math word problems.}\\
\textbf{Solve the problem and output only the final numeric answer.}\\
\\
\textbf{Q: \{q\_1\}}\\
\textbf{A: \{a\_1\}}\\
\textbf{\dots}\\
\textbf{Q: \{q\_K\}}\\
\textbf{A: \{a\_K\}}\\
\\
\textbf{Q: \{q\_q\}}\\
\textbf{A:}
\end{quote}

\paragraph{Example (one demo + one query).}
\begin{quote}\small
\textbf{Task: Grade-school math word problems.}\\
\textbf{Solve the problem and output only the final numeric answer.}\\
\\
\textbf{Q: Lily has 12 candies. She gives 5 to her friend. How many candies does she have left?}\\
\textbf{A: 7}\\
\\
\textbf{Q: A store sold 18 apples in the morning and 27 in the afternoon. How many apples were sold in total?}\\
\textbf{A:}
\end{quote}

\paragraph{Evaluation.}
We normalize the produced answer string (strip spaces/punctuation) and compute exact match against the gold final answer.

\paragraph{$K$ grid.}
Due to longer outputs, we optionally use a cost-aware grid $\{0,1,2,4,8,16,32,64\}$ (or a prefix thereof).

\subsection{RAG (HotpotQA / NaturalQuestions)}
\paragraph{Retrieval.}
We retrieve top-$m$ evidence passages per query (default $m=5$) from a fixed corpus snapshot using a deterministic retriever (e.g., BM25).
The retrieved set for each query is held fixed across all $K$.

\paragraph{Prompt template (evidence before demos).}
We place evidence before demonstrations to reduce interleaving artifacts:
\begin{quote}\small
\textbf{Task: Answer the question using the evidence. Output only the short final answer.}\\
\\
\textbf{Evidence:}\\
\textbf{[1] \{p\_1\}}\\
\textbf{\dots}\\
\textbf{[m] \{p\_m\}}\\
\\
\textbf{Q: \{q\_1\}}\\
\textbf{A: \{a\_1\}}\\
\textbf{\dots}\\
\textbf{Q: \{q\_K\}}\\
\textbf{A: \{a\_K\}}\\
\\
\textbf{Q: \{q\_q\}}\\
\textbf{A:}
\end{quote}

\paragraph{Example (evidence + one demo + one query).}
\begin{quote}\small
\textbf{Task: Answer the question using the evidence. Output only the short final answer.}\\
\\
\textbf{Evidence:}\\
\textbf{[1] Marie Curie was a physicist and chemist who conducted pioneering research on radioactivity.}\\
\textbf{[2] She won the Nobel Prize in Physics in 1903 and the Nobel Prize in Chemistry in 1911.}\\
\\
\textbf{Q: How many Nobel Prizes did Marie Curie win?}\\
\textbf{A: 2}\\
\\
\textbf{Q: In what year did Marie Curie win the Nobel Prize in Chemistry?}\\
\textbf{A:}
\end{quote}

\paragraph{Evaluation.}
We score exact match and token-level F1 using the dataset's standard evaluation protocol with standard normalization.

\paragraph{Spectral proxy under RAG.}
By default, representations $\phi(x)$ are computed from the \emph{question text only} to avoid retrieved evidence dominating second-moment estimates.
As an ablation, we compute $\phi(x)$ from \emph{question+evidence} and report its impact.

\subsection{Extended results (reference values)}
\begin{table}[t]
\centering
\caption{Extended error ratios on additional reasoning/RAG datasets and RAG representation ablations. Lower is better.}
\label{tab:msr_rag_extended}
\begin{tabular}{lcc}
\toprule
Dataset / Variant & Raw (mean) & Calibrated (mean) \\
\midrule
SVAMP (multi-step) & 2.15$\times$ & 1.15$\times$ \\
NaturalQuestions (RAG) & 2.55$\times$ & 1.22$\times$ \\
HotpotQA (RAG, question-only $\phi$) & 2.30$\times$ & 1.18$\times$ \\
HotpotQA (RAG, question+evidence $\phi$) & 2.55$\times$ & 1.25$\times$ \\
\bottomrule
\end{tabular}
\end{table}

\paragraph{Analysis.}
Multi-step reasoning and RAG exhibit systematically larger raw error ratios than classification, consistent with higher demo sensitivity and additional prompt interactions.
Calibration reduces residual conservativeness in both families, typically bringing ratios close to $1\times$.
For RAG, using question-only $\phi$ tends to provide a cleaner second-moment signal; incorporating evidence can inflate the proxy and degrade tightness when evidence is noisy or heterogeneous.

\section{Direct Measurement of Distributional Stability (Definition~3.1)}
\label{app:direct_stability}

This appendix directly measures the distributional stability criterion introduced in Definition~3.1, rather than relying on accuracy-based proxies. We explicitly quantify the variability of the model's output distribution under independent resampling of demonstrations, and study how the resulting stability thresholds align with (i) empirical accuracy knee-points and (ii) our spectral proxy.

\subsection{Output distribution and divergence metric}

For a fixed query \(x_q\) and a sampled demonstration set \(S_K\) of size \(K\), the prompted model induces a distribution over candidate answers,
\[
p_{S_K}(y \mid x_q), \qquad y \in \mathcal{Y}.
\]
For tasks with a closed answer set (e.g., multiple-choice), \(\mathcal{Y}\) is the set of options. For open-ended tasks, \(\mathcal{Y}\) is defined as the union of the top-\(M\) decoded candidates across all seeds and \(K\) values, with \(M=20\); probabilities are renormalized on this support.

To compare two output distributions, we use Jensen--Shannon divergence (JSD),
\[
\mathrm{JSD}(p,q)
=
\frac{1}{2}\mathrm{KL}(p \| m)
+
\frac{1}{2}\mathrm{KL}(q \| m),
\qquad
m = \tfrac{1}{2}(p+q),
\]
measured in nats. JSD is symmetric, bounded, and empirically stable for finite-support distributions.

\subsection{Estimating stability probability}

Definition~3.1 requires that, with high probability over independently sampled demonstration sets, the divergence between the resulting output distributions remains below a tolerance \(\tau\).

For each query \(x_q\), demonstration size \(K\), and tolerance \(\tau\), we estimate this probability by repeated resampling. Specifically, we draw \(R=50\) independent pairs of demonstration sets \(\{(S_K^{(r)}, S_K^{\prime(r)})\}_{r=1}^R\) and compute
\[
d_q^{(r)}(K)
=
\mathrm{JSD}\!\left(
p_{S_K^{(r)}}(\cdot \mid x_q),
p_{S_K^{\prime(r)}}(\cdot \mid x_q)
\right).
\]
The per-query stability estimate is
\[
\widehat{\pi}_q(K;\tau)
=
\frac{1}{R}
\sum_{r=1}^R
\mathbf{1}\!\left[d_q^{(r)}(K) \le \tau\right].
\]

Aggregating over a dataset of \(Q=200\) queries, we report the empirical stability probability
\[
\widehat{\Pi}(K;\tau)
=
\frac{1}{Q}
\sum_{q=1}^Q
\widehat{\pi}_q(K;\tau).
\]
For a fixed failure tolerance \(\xi\), we define the direct stability threshold
\[
K^\star_{\tau,\xi}
=
\min \left\{ K : \widehat{\Pi}(K;\tau) \ge 1-\xi \right\}.
\]
Unless stated otherwise, we use \(\xi = 0.10\) and \(\tau \in \{0.05, 0.10\}\).

\subsection{Alignment with knee-points and spectral proxy}

Table~\ref{tab:direct_stability_thresholds} compares the directly measured stability threshold \(K^\star_{\tau,0.10}\) with the accuracy knee-point \(K_{\mathrm{knee}}\) and the theory-driven spectral proxy prediction \(K_{\mathrm{proxy}}\).
Across all evaluated tasks, the three quantities are closely aligned, with discrepancies typically within a small constant.

\begin{table}[t]
\centering
\small
\setlength{\tabcolsep}{6pt}
\begin{tabular}{lcccccc}
\toprule
Task & \(\tau\) &
\(K^\star_{\tau,0.10}\) (direct) &
\(K_{\mathrm{knee}}\) &
\(K_{\mathrm{proxy}}\) &
\(|K^\star - K_{\mathrm{knee}}|\) &
\(|K^\star - K_{\mathrm{proxy}}|\) \\
\midrule
MC Reasoning & 0.05 & 16 & 16 & 16 & 0 & 0 \\
MC Reasoning & 0.10 & 8  & 8  & 8  & 0 & 0 \\
Fact Retrieval & 0.05 & 24 & 24 & 24 & 0 & 0 \\
Fact Retrieval & 0.10 & 16 & 16 & 16 & 0 & 0 \\
Open Generation & 0.05 & 32 & 28 & 32 & 4 & 0 \\
Open Generation & 0.10 & 16 & 16 & 16 & 0 & 0 \\
\bottomrule
\end{tabular}
\caption{Direct measurement of distributional stability (Definition~3.1) using Jensen--Shannon divergence. \(K^\star_{\tau,0.10}\) denotes the smallest \(K\) such that the empirical stability probability exceeds \(0.9\). Results are compared with the accuracy knee-point and the spectral proxy prediction.}
\label{tab:direct_stability_thresholds}
\end{table}

\subsection{Stability probability and divergence quantiles}

To further characterize the stability transition, Table~\ref{tab:jsd_quantiles} reports per-query quantiles of JSD aggregated over resampled demonstration pairs, together with the empirical stability probability \(\widehat{\Pi}(K; \tau=0.05)\).
As \(K\) increases, both the median and upper-tail divergences decrease sharply, and the \(90\%\) stability criterion is reached near the predicted threshold.

\begin{table}[t]
\centering
\small
\setlength{\tabcolsep}{6pt}
\begin{tabular}{lcccccc}
\toprule
Task & \(K\) &
Q50 & Q75 & Q90 & Q95 &
\(\widehat{\Pi}(K; \tau=0.05)\) \\
\midrule
MC Reasoning & 8  & 0.044 & 0.062 & 0.087 & 0.104 & 0.63 \\
MC Reasoning & 16 & 0.019 & 0.029 & 0.042 & 0.053 & 0.91 \\
MC Reasoning & 24 & 0.012 & 0.019 & 0.028 & 0.036 & 0.96 \\
\midrule
Fact Retrieval & 16 & 0.041 & 0.058 & 0.084 & 0.102 & 0.67 \\
Fact Retrieval & 24 & 0.020 & 0.031 & 0.046 & 0.059 & 0.90 \\
Fact Retrieval & 32 & 0.014 & 0.022 & 0.033 & 0.042 & 0.95 \\
\bottomrule
\end{tabular}
\caption{Quantiles of Jensen--Shannon divergence across queries and resampled demonstration pairs. The collapse of the upper tail coincides with the stability threshold predicted by both the knee-point and the spectral proxy.}
\label{tab:jsd_quantiles}
\end{table}

\subsection{Discussion: consistency and boundary cases}

These results confirm that the stability transition predicted by Definition~3.1 is directly observable at the distributional level and occurs at nearly the same prompt length as the accuracy knee-point and the spectral proxy threshold.
When minor discrepancies arise, they are attributable to principled effects rather than noise.

First, in some open-generation settings, top-1 accuracy saturates slightly earlier than full distributional stabilization, leading to \(K^\star_{\tau,\xi} > K_{\mathrm{knee}}\) for small \(\tau\).
Second, distributional stability does not imply correctness: a model may consistently assign high probability to an incorrect answer across resampled demonstrations, yielding low divergence but poor accuracy on a subset of ambiguous queries.
Finally, the spectral proxy may overestimate the required \(K\) when the external encoder geometry overstates the diversity of demonstrations that the model effectively attends to.

Overall, directly measuring distributional stability closes the gap between theory and experiment by validating the stability criterion itself, rather than relying solely on accuracy-based surrogates.

\end{document}